\documentclass{article} 
\usepackage{iclr2026_conference,times}

\usepackage{amsmath,amsfonts,bm}

\def\eqref#1{equation~\ref{#1}}

\def\1{\bm{1}}

\DeclareMathAlphabet{\mathsfit}{\encodingdefault}{\sfdefault}{m}{sl}
\SetMathAlphabet{\mathsfit}{bold}{\encodingdefault}{\sfdefault}{bx}{n}

\usepackage{hyperref}
\usepackage{url}
\usepackage{graphicx}
\usepackage{xcolor}
\usepackage{multirow}
\usepackage[table,xcdraw,dvipsnames]{xcolor}

\usepackage{tcolorbox}
\tcbuselibrary{listingsutf8}
\usepackage{xcolor}
\newtcblisting{mycodebox}{
  colback=gray!10,    
  colframe=gray!40,   
  listing only,
  listing options={
    basicstyle=\ttfamily\footnotesize,
    language=Python,
    showstringspaces=false,
    breaklines=true,
  },
  left=0pt,
  right=0pt,
  top=2pt,
  bottom=2pt,
  boxsep=4pt,
  arc=4pt,
  enhanced,
}

\usepackage{verbatimbox}
\usepackage{siunitx,amsmath,amssymb}

\usepackage{xspace}
\usepackage{mdwlist}
\usepackage{tabularx}
\usepackage{booktabs}
\usepackage{colortbl}

\newcommand{\rparagraph}[1]{\vspace{1.4mm}\noindent\textbf{#1.}}

\newcommand{\leanfire}{\textsc{Lean-FIRe}\xspace}

\newcommand{\deepseek}{DeepSeek-V3.1\xspace}
\newcommand{\gpt}{GPT-4.1\xspace}
\newcommand{\claude}{Claude-4-Opus\xspace}

\newcommand{\ConjectureBench}{ConjectureBench\xspace}
\newcommand{\ConjectureJudge}{ConJudge\xspace}

\newcolumntype{Y}{>{\centering\arraybackslash}X}

\newtcolorbox{myleanbox}{colback=orange!5!white, colframe=orange!25!white, fonttitle=, coltitle=black, title=Lean 4}

\usepackage{listings,amssymb, color}
\usepackage[normalem]{ulem}
\definecolor{keywordcolor}{rgb}{0.7, 0.1, 0.1}   
\definecolor{tacticcolor}{rgb}{0.0, 0.1, 0.6}    
\definecolor{commentcolor}{rgb}{0.4, 0.4, 0.4}   
\definecolor{symbolcolor}{rgb}{0.0, 0.1, 0.6}    
\definecolor{sortcolor}{rgb}{0.1, 0.5, 0.1}      
\definecolor{attributecolor}{rgb}{0.7, 0.1, 0.1} 
\definecolor{Figure_blue}{HTML}{6C8EBF} 

\title{Conjecturing: An Overlooked Step in Formal Mathematical Reasoning}

\author{
    Jasivan Alex Sivakumar\thanks{Work conducted during an internship at Huawei Noah's Ark Lab, London.} \,\textsuperscript{\normalfont1}, Philipp Borchert\textsuperscript{\normalfont2}, Ronald Cardenas\textsuperscript{\normalfont2}, Gerasimos Lampouras\textsuperscript{\normalfont2} \\
    \textsuperscript{1}University of Sheffield, UK \\
    \textsuperscript{2}Huawei Noah's Ark Lab, London, UK \\
    \texttt{jasivakumar1@sheffield.ac.uk}, \texttt{philipp.borchert@h-partners.com},\\ \texttt{ronald.cardenas.acosta@h-partners.com}, \texttt{gerasimos.lampouras@huawei.com} \\
}

\iclrfinalcopy 
\begin{document}

\maketitle

\begin{abstract}

Autoformalisation, the task of expressing informal mathematical statements in formal language, is often viewed as a direct translation process. This, however, disregards a critical preceding step: conjecturing. Many mathematical problems cannot be formalised directly without first conjecturing a conclusion such as an explicit answer, or a specific bound. Since Large Language Models (LLMs) already struggle with autoformalisation, and the evaluation of their conjecturing ability is limited and often entangled within autoformalisation or proof, it is particularly challenging to understand its effect.
To address this gap, we augment existing datasets to create \ConjectureBench, and redesign the evaluation framework and metric specifically to measure the conjecturing capabilities of LLMs both as a distinct task and within the autoformalisation pipeline. Our evaluation of foundational models, including \gpt and \deepseek, reveals that their autoformalisation performance is substantially overestimated when the conjecture is accounted for during evaluation. However, the conjecture should not be assumed to be provided.
We design an inference-time method, \leanfire to improve conjecturing and autoformalisation, which, to the best of our knowledge, achieves the first successful end-to-end autoformalisation of 13 PutnamBench problems with \gpt and 7 with \deepseek. We demonstrate that while LLMs possess the requisite knowledge to generate accurate conjectures, improving autoformalisation performance requires treating conjecturing as an independent task, and investigating further how to correctly integrate it within autoformalisation. Finally, we provide forward-looking guidance to steer future research toward improving conjecturing, an overlooked step of formal mathematical reasoning.

\end{abstract}

\section{Introduction}

Natural language reasoning with Large Language Models (LLMs) has emerged as a powerful tool for solving complex mathematical problems. Its effectiveness is highlighted by recent breakthroughs, such as AI systems from OpenAI and Google solving five of six problems from the 2025 International Mathematics Olympiad (IMO) using natural language \citep{Gemini-2025-IMOs}. The critical caveat is that these informal solutions require validation by expert mathematicians, a process that is prone to human error and lack scalability \citep{Gouezel-Shchur-2019-lean_error}. Proof assistants like Isabelle \citep{wenzel_etal_2008_isabelle} and Lean \citep{moura2021lean} provide a path toward automated verification at scale through formal reasoning. Their power was demonstrated when AlphaProof solved three of the six 2024 IMO problems by generating formal proofs \citep{AlphaProof2024} and reiterated in 2025 with SeedProver~\citep{chen2025seedproverdeepbroadreasoning} equaling OpenAI and Google's performance. Yet benchmarks such as PutnamBench remain difficult, with the best open-source models achieving a correct proof rate of only 13.1\% at the time of writing \citep{putnam_tsoukalas_etal_2024}.

A central bottleneck is \emph{autoformalisation}, the task of automatically expressing informal mathematics into a precise formal language \citep{szegedy2020position}. On undergraduate-level problems from the ProofNet benchmark \citep{azerbayev2023proofnet}, the current state-of-the-art performance is only 31.28\% \citep{liu2025rethinking}.
Moreover, the fact that state-of-the-art systems like AlphaProof are provided with human-annotated formalisations, rather than the natural language problems, suggests that an end-to-end approach remains challenging. Autoformalisation is non-trivial, as even highly skilled human experts can take over eight hours to formalise a single IMO problem \citep{liu2505combibench}. Improving autoformalisation would therefore be transformative, not only by providing a systematic way to validate informal reasoning but also by enabling the synthesis of new data at scale to improve automated provers themselves.

Autoformalisation is difficult for two interrelated reasons: faithfulness and conjecturing. Without a ground truth formalisation\footnote{In this work, we always assume existence of a ground truth formalisation.}, it can be difficult to judge whether the autoformalised statement truly reflect the intent expressed by the natural language problem~\citep{yang2025position_formal_mathematical_reasoning}.
Humans generally describe problems in an informal manner, often obfuscated through real world objects and situations. To formalise these, LLMs need to connect world knowledge with abstract mathematical concepts, which increases the complexity of the task \citep{yang2025position_formal_mathematical_reasoning}.

Secondly, a conjecture, a mathematical conclusion such as an explicit answer, bound, or proposition, is required for formalisation. The nature of the conjecture shapes the autoformalisation, without which proving stalls. To circumvent conjecturing during autoformalisation, one may insert a placeholder, but it must ultimately be replaced with a valid solution for a complete proof. Most current systems implicitly treat conjecturing as part of the proof search \citep{sun_etal_2025_enumerate-conjecture-prove} by proposing a solution and validating it when a verified proof is generated. However, using a proof as self-verification of the conjecture comes with an important caveat; it does not guarantee completeness. For example, solving $x^2 - 4x = 0$ by conjecturing $x = 0$ yields a valid and verifiable yet incomplete solution, as $x = 4$ is also a valid root. This highlights that conjecturing and proving draw on distinct skills. Conjecturing relies on intuition, pattern recognition, and heuristic testing, whereas proving requires the rigorous application of tactics \citep{Fernández-León28052021_proofVSconjecture}.

\begin{figure}[]
    \centering
    \includegraphics[width=\linewidth]{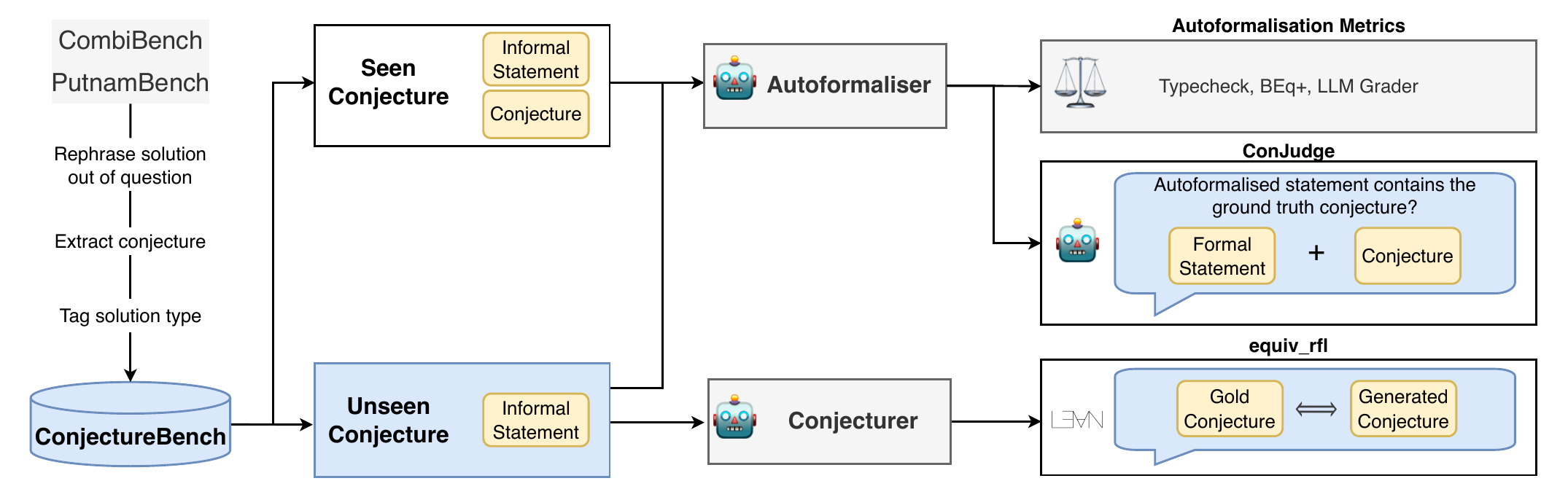}
    \vspace{0.2cm}
    \caption{End-to-end evaluation pipeline for conjecturing and autoformalisation, including a ``seen'' setting (conjecture provided) and a more realistic ``unseen'' setting (conjecture must be inferred). Our contributions, highlighted in \textcolor{Figure_blue}{blue}, introduce \ConjectureBench, ``unseen'' evaluation, and two corresponding metrics: \ConjectureJudge for assessing conjecturing during autoformalisation and \texttt{equiv\_rfl} for standalone conjecture generation.}
    \label{fig: pipeline}
    \vspace{-0.2cm}
\end{figure}

To address the overlooked role of conjecturing in formal mathematical reasoning, we measure the conjecturing capability by introducing \ConjectureBench, a new dataset designed to evaluate the conjecturing performance of LLMs. We develop two novel metrics: \ConjectureJudge, a metric that uses an LLM-as-a-Judge \citep{zheng2023llm-as-a-judge} to assess conjecture presence within the autoformalisation, and \texttt{equiv\_rfl}, a metric that uses Lean tactics to check for definitional equivalence in standalone conjecture generation as illustrated in Figure~\ref{fig: pipeline}. Our evaluation of foundational LLMs, including \gpt and \deepseek, on \ConjectureBench reveals that autoformalisation performance is substantially overestimated when the conjecturing step is assumed to be provided.

To test the hypothesis that this performance gap stems from a failure in reasoning rather than a lack of mathematical and world knowledge, we propose a novel inference-time method \textbf{Lean F}ormal-\textbf{I}nformal \textbf{Re}asoning (\leanfire). This approach guides the model by interleaving Chain-of-Thought (CoT) reasoning in natural language with Lean-of-Thought (LoT) steps in formal language, helping it to better connect informal reasoning with formal mathematics.
We show that \leanfire leads to significant improvements, confirming our hypothesis. While end-to-end autoformalisation remains low, our method achieves the first successful autoformalisation of 13 new PutnamBench ``no-answer'' problems. More specifically, \leanfire improves conjecturing performance on our \ConjectureJudge metric by an average of 29.1\% for \gpt and 14.0\% for \deepseek. These results provide strong evidence that the models' primary limitation is not a lack of requisite knowledge, but rather the need for targeted methods to unlock their ability to conjecture effectively. Lastly, through manual analysis, we further identify two practical challenges: dataset contamination and the need for new definitions, functions, and lemmata to support autoformalisation.

Our contributions are as follows: (1) we introduce \ConjectureBench\footnote{The dataset and code are available at \url{https://github.com/huawei-noah/ConjectureBench}}, the first benchmark evaluating conjecture capabilities, (2) we propose two complementary metrics, \ConjectureJudge and \texttt{equiv\_rfl}, to systematically assess thse capabilities, and (3) we develop \leanfire, an inference-time method to improve both autoformalisation and conjecturing.

\section{Preliminary}
\begin{table}[t!]
\centering
\footnotesize  
\[
\begin{array}{ccc|c}
\toprule
\textbf{Hypothesis} & \Rightarrow & \textbf{Conclusion} & \textbf{Type of solution} \\
\midrule
x+4=0 & \Rightarrow & x=-4 & \text{Numerical} \\
\midrule
x^2-a=0 & \Rightarrow &
    x \in \{\sqrt{a}, -\sqrt{a}\}
    & \text{Algebraic} \\
\midrule
\cos(x) = x & \Rightarrow & \exists x \text{ s.t. } \cos(x)=x & \text{True but no closed-form solution} \\
\bottomrule
\end{array}
\]
\caption{Examples of mathematical statements paired with different solution types.}
\label{tab: type of solutions}
\end{table}

In mathematics, a theorem is a statement for which a proof establishes a conclusion from a set of hypotheses. When such a proof is not yet known, the statement is referred to as a conjecture \citep{math_defintions}. A conjecture proposes a possible conclusion often expressed as an abstract object that may or may not admit a closed-form representation such as an algebraic formula or a numerical answer, see Table~\ref{tab: type of solutions}.
In formal mathematics, autoformalisation is a necessary stage prior to using a prover or proof assistant, as these systems require formal statements as inputs. Conjecturing is the task of generating candidate solutions for well-posed problems \citep{sun_etal_2025_enumerate-conjecture-prove}.

Current formal mathematics datasets largely fall into two categories. The first type assumes that a solution is already known and only requires the corresponding proof given a gold formalised statement. The second type requires the discovery of a solution before or while a proof is constructed. For this latter class, the initial step is to generate a candidate solution. Without such a conjecture, formalisation cannot proceed. This holds in Lean 4, a more permissive formal mathematics language; the compiler cannot verify whether the object types are consistent (Typecheck) in an incomplete statement. 

\begin{myleanbox}
\begin{lstlisting}
theorem quad_roots: {x : ℝ | x^2 - 4*x = 0} (*@\sout{= conjecture}@*) := sorry
\end{lstlisting}
\end{myleanbox}
\vspace{0.5cm}

In the above \texttt{quad\_roots} example, the formal statement for ``What are the real roots of $x^2-4x$?", erasing \texttt{conjecture} reduces the statement to a set of hypotheses with no conclusion, leaving nothing to prove. A quick fix is to put a placeholder, \texttt{conjecture}, for which Lean 4 has been forced to assume the correct type. When the solution is known, it could be integrated directly into the formal statement. But deriving it in the first place is challenging. If generated during the proving stage, the formal language system can self-verify whether the conjecture is valid. However, the validity of a conjecture does not equate to a \emph{complete} conjecture or a valid solution to the informal statement. Three valid and proof verifiable conjectures are:

\begin{table}[h!]
\begin{myleanbox}
\centering
\begin{tabular}{lcr}
\begin{lstlisting}
abbrev conjecture_1:
  Set ℝ := {0}
\end{lstlisting}
&
\begin{lstlisting}
abbrev conjecture_2:
  Set ℝ := {4}
\end{lstlisting}
&
\begin{lstlisting}
abbrev conjecture_3:
  Set ℝ := {0, 4}
\end{lstlisting}
\end{tabular}
\end{myleanbox}
\end{table}

However, only \texttt{conjecture\_3} is a complete answer.
In fact, natural language can frame a problem in a way that feels more intuitive and human-friendly. For example,
\emph{``How many people must be in a group for at least two of them to be born in the same month?''}, this question is easier to reason about using everyday knowledge than its more formal counterpart: determining the smallest domain size for which there exist no injective function into a set of 12 elements. Therefore, autoformalisation being closer to the natural language statement allows for broader possibility of generating conjectures. Finally, when tackling unsolved problems, the solution is not given in advance making conjecture generation an essential step in the formal reasoning process. Therefore, this motivates our exploration of conjecturing as an integral, yet overlooked step in formal mathematical reasoning.

\section{Methodology}\label{sec: methodology}
\subsection{\ConjectureBench Dataset}
Two recent datasets are designed with conjecturing in mind: PutnamBench \citep{putnam_tsoukalas_etal_2024} factors out the solution from the problem statement, forcing models to generate the conjecture itself, while CombiBench \citep{liu2505combibench} introduces a benchmark with and without the solution to further encourage conjecture generation. To elaborate, PutnamBench is a benchmark of 640 paired informal and formal statements from the William Lowell Putnam Mathematical Competition. The benchmark and its leaderboard primarily emphasise proof generation, both when solutions are provided and when they are withheld. The evaluation of statements without answers is only feasible for 355 of the problems. Similarly, CombiBench adopts the same design where possible, with 100 combinatorics problems ranging from textbook exercises to IMO questions. However, 55 questions include the conjecture within their informal statement.

\begin{table}[h!]
\centering
\resizebox{\textwidth}{!}{\begin{tabular}{c|c|c|c}
\toprule
\textbf{Original with integrated solution}                                                                      & \textbf{Reworded to seek a solution}                                                                                      & \textbf{Type of solution} & \textbf{Distribution}                                \\ \midrule
\begin{tabular}[c]{@{}c@{}}Show that there are at least\\ 1991 red points in the plane.\end{tabular}   & \begin{tabular}[c]{@{}c@{}}What is the minimum number\\ of red points in the plane?\end{tabular}                  & Numerical        & \begin{tabular}[c]{@{}c@{}}39.0\%\\ (178)\end{tabular} \\ \midrule
\begin{tabular}[c]{@{}c@{}}Prove that there are at most\\ $2n-1$ subsets in the collection.\end{tabular} & \begin{tabular}[c]{@{}c@{}}What is the maximum number of subsets\\ that can be in such a collection?\end{tabular} & Algebraic        & \begin{tabular}[c]{@{}c@{}}36.1\%\\ (165)\end{tabular} \\ \midrule
\begin{tabular}[c]{@{}c@{}}Prove that B = \{0, 3, 4, 9, 11\}\\ is a difference set in $Z_{21}$.\end{tabular}    & \begin{tabular}[c]{@{}c@{}}Prove or disprove that B = \{0, 3, 4, 9, 11\}\\ is a difference set in $Z_{21}$.\end{tabular}   & Proof            & \begin{tabular}[c]{@{}c@{}}24.9\%\\ (114)\end{tabular} \\
\bottomrule
\end{tabular}}
\caption{Examples of how proof questions are reformulated into the three solution types considered, along with the distribution of these types in \ConjectureBench.}
\label{tab: reword questions}
\end{table}

To adapt both datasets to evaluate conjecturing, we first annotated all 355 PutnamBench problems and 102 CombiBench problems (splitting multi-part questions into separate items) to ensure that no conjecture appear directly in the problem statements. For proof-based questions, where the conclusion is already embedded, we rephrased them into equivalent tasks requiring either a numerical or algebraic solution. When rewording is not feasible, we instead reformulate the problem into a binary classification task, requiring the model to decide whether the statement is true or false. Examples of these reformulations, as well as the distribution across our new combined dataset, \ConjectureBench, are provided in Table~\ref{tab: reword questions}. We finally separate the conjecture from the formal statement, retaining it only in the ``seen’’ setting as illustrated in Figure~\ref{fig: pipeline}. This design choice ensures that our full dataset of 457 paired informal–formal statements can be used consistently across both, ``seen'' and ``unseen'' settings, enabling a more accurate evaluation of conjecturing.

This evaluation framework offers several advantages. It allows us to assess whether current LLMs are capable of generating accurate conjectures while autoformalising, but also to evaluate models' raw conjecturing capability. It also enables a detailed analysis of which types of conjectures present particular challenges for existing models. The results of this benchmark provide a foundation to investigate whether improvements in conjecturing arise naturally from enhanced autoformalisation, or if alternative approaches, such as new data or reasoning approaches, are necessary.

\begin{figure}[t]
    \centering
    \includegraphics[width=1\linewidth]{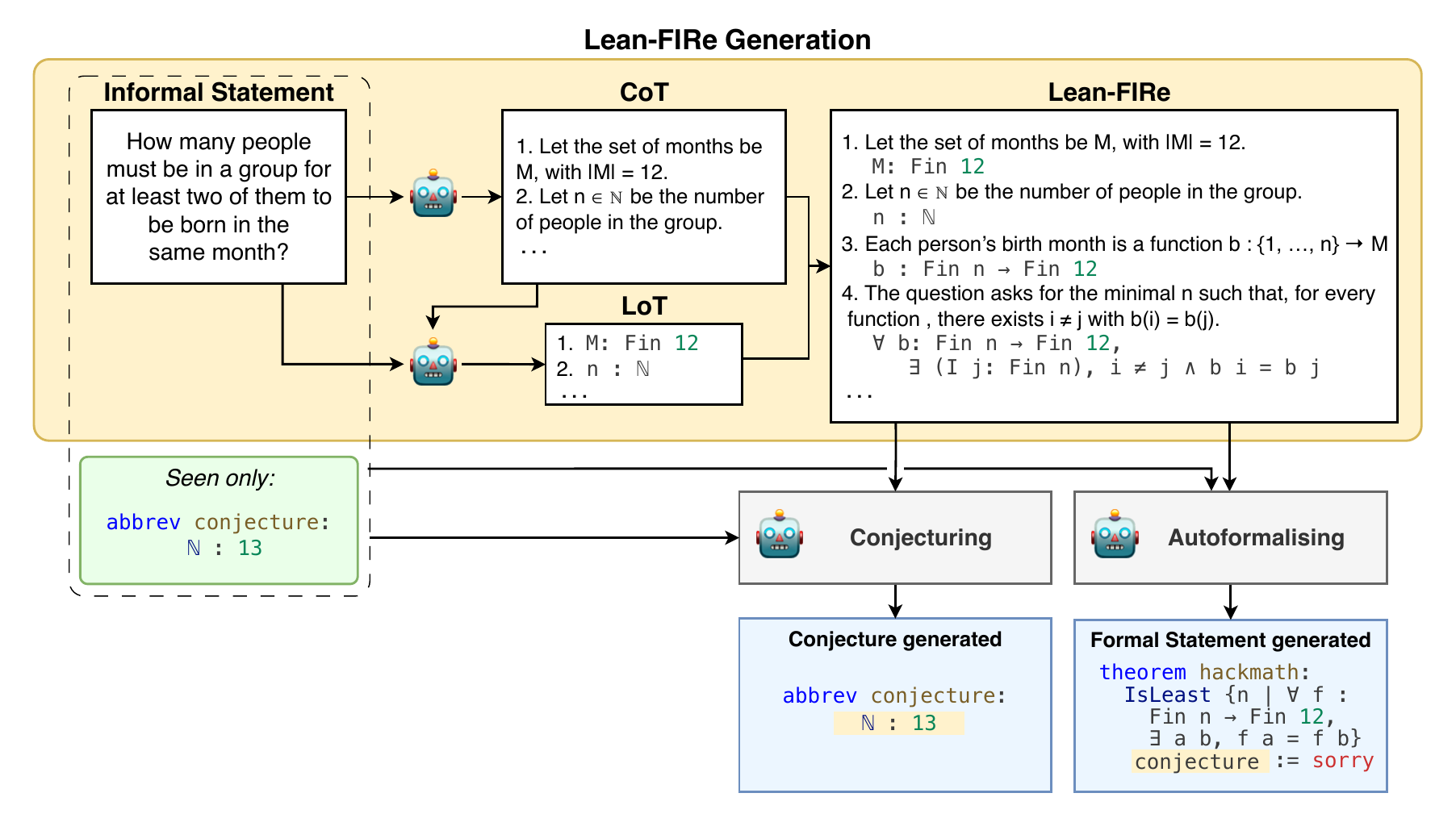}
    \caption{Illustration of \leanfire construction within the overall pipeline for generating autoformalisations and conjectures, where the conjecture in the green box is provided only in the ``seen'' setting, and CoT and LoT stand for chain/lean-of-thought.}
    \label{fig:leanfire}
    \vspace{-0.4cm}
\end{figure}
\subsection{Conjecturing Tasks}
We evaluate performance across two distinct tasks designed to assess conjecture-driven reasoning as illustrated in Figure~\ref{fig:leanfire}. 
The primary task is \emph{autoformalisation}, which we evaluate in two settings. In the ``seen'' setting, the model is provided with the informal statement and the correct conjecture formatted in Lean 4. The task is to produce a formal statement that correctly incorporates the provided conjecture. In the ``unseen'' setting, the model is provided only with the informal statement and must deduce and incorporate the conjecture directly into the final formalisation.

The second task is \emph{standalone conjecture generation}, where we isolate conjecturing performance entirely from the complexity of full autoformalisation. Here, the model is given only the informal statement and is instructed to generate the conjectured solution as a concise Lean 4 statement.

\subsection{Metrics}\label{sec: metrics}
To evaluate conjecturing performance during autoformalisation, we propose \ConjectureJudge, an LLM-as-a-judge framework \citep{zheng2023llm-as-a-judge}. Its purpose is to determine whether the problem's gold solution is reasonably and correctly incorporated as a conjecture within the final autoformalised statement. To do this, \ConjectureJudge is provided with the generated formalisation, the gold conjecture, and the gold formalisation to demonstrate the intended context and role of the conjecture. For instance, if the correct conjecture is the integer $2$, the judge would reject a formalisation where $2$ appears incorrectly as a power or a subscript.
To tune \ConjectureJudge, we carry out a human annotation of 100 randomly sampled autoformalisation generations (Appendix~\ref{app: ConJudge}), classifying whether the solution was correctly incorporated into the formal statement.

For standalone conjecture generation, we created \texttt{equiv\_rfl}, which evaluates definitional equivalence between the generated and gold conjecture based on tactic \texttt{rfl} (Appx.~\ref{app: equiv_rfl}). This provides a rigorous, formal measure of whether the model can produce the correct solution in isolation.

\subsection{Lean-guided Formal-Informal Reasoning (\leanfire)}\label{sec: Inference-time strategies}
To test the hypothesis that the performance gap in conjecturing stems from a failure in reasoning rather than a lack of knowledge, we propose \leanfire, a novel inference-time method designed to better structure the model's reasoning process. The goal is to distil the LLM's latent parametric mathematical knowledge at test-time by combining both informal and formal reasoning. 
As illustrated in Figure~\ref{fig:leanfire}, the \leanfire method is built as a two-stage hybrid reasoning process that integrates informal problem decomposition with formal code generation by means of interleaved Chain-of-Thought (CoT) with Lean-of-Thought (LoT) prompting. We leverage the LLM's ability in informal mathematical reasoning to first generate a potential conjecture and outline the overall structure of the formalisation.
First, a complete CoT trace is generated in natural language from the informal problem statement. The CoT is designed to break down the problem, identify key mathematical objects, and articulate the reasoning entirely in natural language. Crucially, this phase is constrained to produce no formal code and avoid stating the final solution. Second, after the informal reasoning trace is completed, a subsequent LLM generates a corresponding LoT step for each informal step. The purpose of the LoT is not to write a comprehensive formal statement, but to translate the abstract concepts from the CoT into precise Lean primitives and syntax. This hybrid approach is motivated in part by prior work, such as \cite{jiang2023draft_sketch_prove}, which has already demonstrated that leveraging both formal and informal language can improve performance in theorem proving.

\rparagraph{Seed Data Annotation}\label{sec: Seed Data Annotation}
This automated generation of CoT and LoT steps is enabled by few-shot examples derived from a small, expert-annotated seed dataset. We created this seed data from five diverse Putnam competition problems, which were annotated by an expert mathematics instructor to produce gold CoTs. The problems were selected to cover a range of mathematical domains (probability, real analysis, linear algebra, abstract algebra, number theory), solution types (as listed in Table~\ref{tab: reword questions}), and conjecture styles, ensuring the exemplars were broadly representative. In some cases, questions were modified to omit parts of the solution, mirroring the annotation process for \ConjectureBench. These five seed problems are detailed in Appendix~\ref{app: seed-questions} and are excluded from our \ConjectureBench evaluation. With these few-shot examples and a set of precise instructions (see Appendix~\ref{app:cot-lot-instructions}), CoT and LoT pairs can be automatically generated for any new problem using only its informal statement as input. In preliminary experiments, we evaluated five LLMs for this task and found that \gpt consistently outperformed its other family models and \claude.

\section{Experimental Setup}
\label{sec:experiments}
\rparagraph{Models} 
We experiment with two foundational autoformalisation models: \gpt~\citep{achiam2023gpt} and \deepseek~\citep{liu2024deepseek}. To measure the impact of our proposed method, we compare the performance of \leanfire against the zero-shot performance of each base model. Additionally, we conduct an ablation study where we remove the few-shot examples from the \leanfire input (w/o FS) to isolate the contribution of the hybrid reasoning approach.

\rparagraph{Metrics} 
We assess performance for all tasks using pass@1 and pass@10, where pass@$k$ indicates that at least one of $k$ independent samples was successful.
For conjecturing, we use two targeted metrics. Conjecturing performance during the full autoformalisation task is assessed with \textbf{\ConjectureJudge}, while standalone conjecture generation is evaluated using \textbf{\texttt{equiv\_rfl}}.

For autoformalisation, we use three complementary metrics: \textbf{Typecheck}, \textbf{BEq+}, and \textbf{LLM Grader}.
\textbf{Typecheck} is a binary measure of syntactic correctness indicating whether the generated Lean code compiles without error.\footnote{\hangindent=1.8em Each instance of \ConjectureBench is provided with the appropriate Mathlib imports and a standardised Lean 4 environment (\texttt{v4.19.0-rc2}) to ensure consistent evaluation.}
For semantic equivalence, we use \textbf{BEq+}, a metric based on a set of Lean tactics that presupposes typechecking and attempts to prove equivalence between the generated and gold formalisations \citep{poiroux2025improvingautoformalizationusingtype}. We should note that while precise, BEq+ can be overly conservative, leading to false negatives on semantically equivalent statements that differ in surface form \citep{liu2025rethinking}.
To capture a broader notion of correctness, we also use \textbf{LLM Grader}, a pipeline that evaluates semantic alignment. First, the gold and generated formalisations are back-translated into natural language using a math LLM.\footnote{We employ InternLM2-Math-Plus-20B~\citep{internlm2}.} A separate judge LLM\footnote{We employ a Qwen3-14B~\citep{yang2025qwen3technicalreport} calibrated against human annotators (see Appendix~\ref{app: ConJudge}).} then evaluates these natural language statements for semantic equivalence.

\section{Results and Discussion}

\subsection{Conjecturing Results}

\begin{table}[h]
\begin{center}
{
\def\arraystretch{1.1}
\setlength{\tabcolsep}{6pt} 
\fontsize{9.5pt}{9.7pt}\selectfont
\centering
\begin{tabularx}{0.85\linewidth}{l l l X X}
\toprule
\textbf{Model} & \textbf{Method} & \textbf{Conjecture} & \textbf{ConJudge@1} & \textbf{ConJudge@10} \\
\midrule
\multirow{6}{*}{\rotatebox[origin=c]{90}{\textbf{\gpt}}} & \multirow{2}{*}{Baseline} & \cellcolor{gray!15}Seen & \cellcolor{gray!15} 78.77 & \cellcolor{gray!15} 98.03 \\
& & Unseen & 26.70 \tiny{${\textcolor{Cerulean}{(-52.07)}}$} & 61.27 \tiny{${\textcolor{Cerulean}{(-36.76)}}$} \\
\cmidrule{2-5}
& \multirow{2}{*}{\leanfire} & \cellcolor{gray!15}Seen & \cellcolor{gray!15} 92.78 & \cellcolor{gray!15} 98.47 \\
& & Unseen & \textbf{55.80} \tiny{${\textcolor{Cerulean}{(-36.98)}}$} & \textbf{85.34} \tiny{${\textcolor{Cerulean}{(-13.13)}}$} \\
\cmidrule{2-5} 
& \multirow{2}{*}{\leanfire w/o FS} & \cellcolor{gray!15}Seen & \cellcolor{gray!15} 77.90 & \cellcolor{gray!15} 96.06 \\
& & Unseen & 28.88 \tiny{${\textcolor{Cerulean}{(-49.02)}}$} & 56.89 \tiny{${\textcolor{Cerulean}{(-39.17)}}$} \\
\midrule
\multirow{6}{*}{\rotatebox[origin=c]{90}{\textbf{\deepseek}}} & \multirow{2}{*}{Baseline} & \cellcolor{gray!15}Seen & \cellcolor{gray!15} 80.31 & \cellcolor{gray!15} 95.84 \\
& & Unseen & 30.63 \tiny{${\textcolor{Cerulean}{(-49.68)}}$} & 58.86 \tiny{${\textcolor{Cerulean}{(-36.98)}}$} \\
\cmidrule{2-5}
& \multirow{2}{*}{\leanfire} & \cellcolor{gray!15}Seen & \cellcolor{gray!15} 81.40 & \cellcolor{gray!15} 97.81 \\
& & Unseen & \textbf{44.64} \tiny{${\textcolor{Cerulean}{(-36.76)}}$} & \textbf{71.55} \tiny{${\textcolor{Cerulean}{(-26.26)}}$} \\
\cmidrule{2-5} 
& \multirow{2}{*}{\leanfire w/o FS} & \cellcolor{gray!15}Seen & \cellcolor{gray!15} 74.62 & \cellcolor{gray!15} 96.72 \\
& & Unseen & 35.01 \tiny{${\textcolor{Cerulean}{(-39.61)}}$} & 56.86 \tiny{${\textcolor{Cerulean}{(-39.83)}}$} \\
\bottomrule
\end{tabularx}
}
\caption{Conjecturing during autoformalisation performance on \ConjectureBench using \ConjectureJudge. Scores are reported at pass@1 and pass@10, with relative differences between ``unseen'' and ``seen'' in brackets. Bold indicates best performance for each model and metric in the ``unseen'' setting.}
\label{tab: llm_as_a_judge_results_delta}
\vspace{-0.3cm}
\end{center}
\end{table}
\begin{table*}[htb!]
\centering
{
\def\arraystretch{1.2}
\setlength{\tabcolsep}{6pt}
\fontsize{9.5pt}{9.7pt}\selectfont
\begin{tabularx}{0.7\linewidth}{l l X X}
\toprule
\textbf{Model} & \textbf{Type of solution} & \textbf{equiv\_rfl@1} & \textbf{equiv\_rfl@10} \\
\midrule
\multirow{4}{*}{\textbf{\gpt}}
 & All      & \phantom{0}3.28 \tiny{${(15/457)}$} & \phantom{0}5.04 \tiny{${(23/457)}$} \\
 & Numerical & \textbf{\phantom{0}5.62} \tiny{${(10/178)}$} & \textbf{\phantom{0}8.99} \tiny{${(16/178)}$}\\
 & Algebraic & \phantom{0}3.03 \tiny{${(\phantom{0}5/165)}$}& \phantom{0}4.24 \tiny{${(\phantom{0}7/165)}$}\\
 & Proof     & \phantom{0}0.00 \tiny{${(\phantom{0}0/114)}$}& \phantom{0}0.00 \tiny{${(\phantom{0}0/114)}$}\\
\midrule
\multirow{4}{*}{\textbf{\deepseek}}
 & All       & \phantom{0}3.72 \tiny{${(17/457)}$}& \phantom{0}5.70 \tiny{${(26/457)}$}\\
 & Numerical & \textbf{\phantom{0}7.30} \tiny{${(13/178)}$}& \textbf{10.67} \tiny{${(19/178)}$}\\
 & Algebraic & \phantom{0}2.42 \tiny{${(\phantom{0}4/165)}$}& \phantom{0}3.64 \tiny{${(\phantom{0}6/165)}$}\\
 & Proof     & \phantom{0}0.00 \tiny{${(\phantom{0}0/114)}$}& \phantom{0}0.88 \tiny{${(\phantom{0}1/114)}$}\\
\bottomrule
\end{tabularx}
}
\caption{Standalone conjecture generation performance across \ConjectureBench broken down by type of solution. Metrics report \texttt{equiv\_rfl} at pass@1 and pass@10, with counts shown over total examples in brackets.}
\label{tab: equiv_rfl_solution_types}
\end{table*}

\rparagraph{Conjecturing During Autoformalisation}
Using the \ConjectureJudge metric, we find that models are more adept at producing the correct conjecture when it is part of a full autoformalisation task. Table~\ref{tab: llm_as_a_judge_results_delta} shows that \leanfire with few-shot examples significantly improves the use of conjectures in both ``seen'' and ``unseen'' settings, boosting \gpt's pass@10 by up to 28\% in the ``unseen'' setting.
However, the large performance drop when few-shot examples are removed (w/o FS) indicates that the hybrid reasoning structure alone does not significantly improve conjecturing. Instead, the few-shot examples, which expose the model to various solution types and map reasoning steps to the correct conjecture format, provide the primary benefit. This suggests that a model's ability to conjecture is less a matter of latent reasoning and more a function of direct exposure, pointing to the need for larger and higher-quality conjecture datasets for training.

\rparagraph{Standalone Conjecture Generation}
As shown in Table~\ref{tab: equiv_rfl_solution_types}, performance on standalone conjecture generation is notably low across all models. While models occasionally produce correct numerical conjectures, they more often generate auxiliary constructs such as definitions or lemmata instead of the conjecture itself. The performance on this task is nearly an order of magnitude lower than for conjecturing during autoformalisation (see Table~\ref{tab: llm_as_a_judge_results_delta}), suggesting that models rely heavily on prior exposure to conjectures already embedded within complete formalised solutions. We observed signs of data contamination in the outputs; for instance, some generations used helper functions like \texttt{IsMagicSquare}, which appear only in the gold formalisation of the benchmark. 

\subsection{Autoformalisation Results}
\begin{table}[h!]
{
\def\arraystretch{1.1}
\setlength{\tabcolsep}{2pt}
\fontsize{7.5pt}{7.7pt}\selectfont
\centering
\begin{tabularx}{\linewidth}{l l l X X X | X X X}
\toprule
\textbf{Model} & \textbf{Method} & \textbf{Conjecture} 
& \textbf{TC@1} & \textbf{BEq+@1} & \textbf{Grader@1} 
& \textbf{TC@10} & \textbf{BEq+@10} & \textbf{Grader@10} \\
\midrule
\multirow{6}{*}{\rotatebox[origin=c]{90}{\textbf{\gpt}}} 
  & \multirow{2}{*}{Baseline} 
    & \cellcolor{gray!15}Seen & \cellcolor{gray!15} 25.38 & \cellcolor{gray!15} \phantom{0}0.00 & \cellcolor{gray!15} \phantom{0}7.22 & \cellcolor{gray!15} 59.52 & \cellcolor{gray!15} \phantom{0}6.78 & \cellcolor{gray!15} 36.32 \\
 & & Unseen & 24.29\tiny{${\textcolor{Cerulean}{(-1.09)}}$} 
               & \phantom{0}0.22\tiny{${\textcolor{Green}{(+0.22)}}$} 
               & \phantom{0}3.50\tiny{${\textcolor{Cerulean}{(-3.72)}}$} 
               & \textbf{51.42}\tiny{${\textcolor{Cerulean}{(-8.10)}}$} 
               & \phantom{0}\textbf{4.38}\tiny{${\textcolor{Cerulean}{(-2.40)}}$} 
               & 20.35\tiny{${\textcolor{Cerulean}{(-15.97)}}$} \\
\cmidrule{2-9}
  & \multirow{2}{*}{\leanfire} 
    & \cellcolor{gray!15}Seen & \cellcolor{gray!15} 31.95 & \cellcolor{gray!15} \phantom{0}3.72 & \cellcolor{gray!15} 11.82 & \cellcolor{gray!15} 50.98 & \cellcolor{gray!15} \phantom{0}6.56 & \cellcolor{gray!15} 43.33 \\
  & & Unseen & 28.01\tiny{${\textcolor{Cerulean}{(-3.94)}}$} 
               & \phantom{0}1.31\tiny{${\textcolor{Cerulean}{(-2.41)}}$} 
               & \phantom{0}4.60\tiny{${\textcolor{Cerulean}{(-7.22)}}$} 
               & 43.76\tiny{${\textcolor{Cerulean}{(-7.22)}}$} 
               & \phantom{0}3.06\tiny{${\textcolor{Cerulean}{(-3.50)}}$} 
               & 22.76\tiny{${\textcolor{Cerulean}{(-20.57)}}$} \\
\cmidrule{3-9} 
  & \leanfire 
    & \cellcolor{gray!15}Seen & \cellcolor{gray!15} 35.89 & \cellcolor{gray!15} \phantom{0}2.84 & \cellcolor{gray!15} \phantom{0}7.66 & \cellcolor{gray!15} 49.02 & \cellcolor{gray!15} \phantom{0}4.60 & \cellcolor{gray!15} 40.04 \\
  & w/o FS & Unseen & \textbf{28.45}\tiny{${\textcolor{Cerulean}{(-7.44)}}$} 
               & \phantom{0}\textbf{2.41}\tiny{${\textcolor{Cerulean}{(-0.43)}}$} 
               & \phantom{0}\textbf{5.69}\tiny{${\textcolor{Cerulean}{(-1.97)}}$} 
               & 42.67\tiny{${\textcolor{Cerulean}{(-6.35)}}$} 
               & \phantom{0}4.16\tiny{${\textcolor{Cerulean}{(-0.44)}}$} 
               & \textbf{23.85}\tiny{${\textcolor{Cerulean}{(-16.19)}}$} \\
\midrule
\multirow{6}{*}{\rotatebox[origin=c]{90}{\textbf{\deepseek}}} 
  & \multirow{2}{*}{Baseline} 
    & \cellcolor{gray!15}Seen & \cellcolor{gray!15} 38.29 & \cellcolor{gray!15} \phantom{0}4.81 & \cellcolor{gray!15} \phantom{0}6.78 & \cellcolor{gray!15} 61.71 & \cellcolor{gray!15} \phantom{0}6.78 & \cellcolor{gray!15} 35.67 \\
  & & Unseen & 33.26\tiny{${\textcolor{Cerulean}{(-5.03)}}$} 
               & \phantom{0}2.63\tiny{${\textcolor{Cerulean}{(-2.18)}}$} 
               & \phantom{0}5.25\tiny{${\textcolor{Cerulean}{(-1.53)}}$} 
               & 54.49\tiny{${\textcolor{Cerulean}{(-7.22)}}$} 
               & \phantom{0}\textbf{5.47}\tiny{${\textcolor{Cerulean}{(-1.31)}}$} 
               & 24.95\tiny{${\textcolor{Cerulean}{(-10.72)}}$} \\
\cmidrule{2-9}
  & \multirow{2}{*}{\leanfire} 
    & \cellcolor{gray!15}Seen & \cellcolor{gray!15} 46.17 & \cellcolor{gray!15} \phantom{0}3.72 & \cellcolor{gray!15} \phantom{0}9.85 & \cellcolor{gray!15} 66.74 & \cellcolor{gray!15} \phantom{0}6.13 & \cellcolor{gray!15} 41.36 \\
  & & Unseen & \textbf{42.89}\tiny{${\textcolor{Cerulean}{(-3.28)}}$} 
               & \phantom{0}2.63\tiny{${\textcolor{Cerulean}{(-1.09)}}$} 
               & \phantom{0}6.13\tiny{${\textcolor{Cerulean}{(-3.72)}}$} 
               & \textbf{59.30}\tiny{${\textcolor{Cerulean}{(-7.44)}}$} 
               & \phantom{0}4.16\tiny{${\textcolor{Cerulean}{(-1.97)}}$} 
               & \textbf{26.91}\tiny{${\textcolor{Cerulean}{(-14.45)}}$} \\
\cmidrule{3-9}
  & \leanfire
    & \cellcolor{gray!15}Seen & \cellcolor{gray!15} 39.82 & \cellcolor{gray!15} \phantom{0}3.50 & \cellcolor{gray!15} \phantom{0}9.41 & \cellcolor{gray!15} 56.24 & \cellcolor{gray!15} \phantom{0}4.16 & \cellcolor{gray!15} 39.39 \\
  & w/o FS & Unseen & 39.61\tiny{${\textcolor{Cerulean}{(-0.21)}}$} 
               & \phantom{0}2.63\tiny{${\textcolor{Cerulean}{(-0.87)}}$} 
               & \phantom{0}\textbf{6.35}\tiny{${\textcolor{Cerulean}{(-3.06)}}$} 
               & 53.83\tiny{${\textcolor{Cerulean}{(-2.41)}}$} 
               & \phantom{0}3.72\tiny{${\textcolor{Cerulean}{(-0.44)}}$} 
               & 23.63\tiny{${\textcolor{Cerulean}{(-15.76)}}$} \\
\bottomrule
\end{tabularx}
}
\caption{Autoformalisation performance of all models and methods (as percentages) on \ConjectureBench across seen and unseen settings. Metrics include TC (Typecheck), BEq+, and Grader (LLM Grader), reported at pass@1 and pass@10. Unseen results show the difference relative to seen performance in brackets. Bold values indicate the best performance for each model and metric in the ``unseen'' setting.}
\label{tab: autoformalisation_results_delta}
\end{table}

Table~\ref{tab: autoformalisation_results_delta} shows that correct end-to-end autoformalisation remains a challenging task, with low success rates even in the ``seen'' setting where the conjecture is provided. Performance is systematically overestimated in this setting, with an average 23.7\% drop in performance when moving from the ``seen'' to the ``unseen'' setting. 
Despite these challenges, \leanfire achieves notable successes. Generating conjectures, as underscored by the PutnamBench ``no-answer'' leaderboard, was considered as a challenge with no successful submissions to date \citep{putnam_tsoukalas_etal_2024}. Yet, even under the strict BEq+ metric, \leanfire enables \gpt to correctly autoformalise \num{13} new PutnamBench problems and \deepseek to solve \num{7}. To our knowledge, these represent the first successful autoformalisations on PutnamBench in a setting where the solution is withheld.

In contrast to its effect on conjecturing, \leanfire's impact on autoformalisation is more nuanced. When comparing across metrics, both models show consistent gains under Typecheck and LLM Grader. Higher Typecheck scores indicate improved syntactic correctness, while better LLM Grader scores point to improved semantic equivalence. Therefore, the limited gains in BEq+ suggest that assembling correct components into a fully equivalent formalisation remains a key bottleneck. For example, in the generated formalisation of \texttt{putnam\_2014\_b2} below, both Typecheck and LLM Grader marked the output as correct, but BEq+ did not due to a subtle error: a misplaced factorial symbol. This highlights the sensitivity of BEq+ and illustrates that even when all components are present, models may fail to assemble them with complete accuracy.

\begin{myleanbox}
\begin{lstlisting}
abbrev conjecture: (fun n : ℕ => (-1)^(n - 1) / ((n - 1)! * n(*@\textbf{!)}@*))

theorem putnam_2014_a2 : ∀ n : ℕ, 0 < n
    → let A : Matrix (Fin n) (Fin n) ℚ := λ i j
    => 1 / (min (i.val + 1) (j.val + 1) : ℚ) in det A
    = ((-1) ^ (n - 1) : ℚ) / ((n - 1)! * n(*@\textbf{)!}@*))
    := sorry
\end{lstlisting}
\end{myleanbox}
\vspace{0.2cm}

In general, the comparison with the baseline reveals no consistent performance benefit. In the ``seen'' setting, few-shot examples are helpful, but in the ``unseen'' setting, they can be detrimental, sometimes wrongly encouraging template solutions where a conjecture is introduced as a separate function and then integrated into the formalisation. This suggests that the mathematical knowledge required for complex autoformalisation including conjecturing is not fully latent in the model's parameters, or that \leanfire, in its current form, fails to consistently extract it.
\leanfire shows a net mean gain of 3.01\% at pass@1 but a slight decline at pass@10, suggesting that the reasoning guidance primarily helps steer the model's token distribution towards correctness, but the effect is diluted when multiple generations are sampled by the increase of the probability of reaching a better distribution. Still, from Table~\ref{tab: autoformalisation_results_delta}, best-of-n sampling roughly doubles improvement under BEq+ and quadruples it under the LLM Grader, indicating that necessary knowledge exists in latent space, but is hard to reliably retrieve.

\section{Related Work}

Several approaches to autoformalisation leverage retrieval or supervised fine-tuning to bootstrap formal reasoning. For example, \citet{liu2025rethinking} incorporate retrieval to ground the translation process, while \citet{lin2025goedel1} train on large corpora containing both human and synthetic annotations derived from the Lean Workbook \citep{ying2024lean_workbook}, exposing the model to a diverse range of formalisation examples. Data-centric strategies, focusing on increasing dataset size or improving data quality, are also common. Some methods employ LLMs-as-a-judge \citep{wang2025kimina}, chain-of-thought (CoT) model scoring \citep{xin2024deepseekprover}, Lean typechecking signals \citep{lu2024processdrivenautoformalizationlean4}, or LLM feedback \citep{peng2025criticlean}. In addition, \citet{sun_etal_2025_enumerate-conjecture-prove} combine typechecking feedback with retrieval within their framework to further enhance autoformalisation performance.

Autoformalisation is also employed in theorem proving: for instance, \citet{jiang2023draft_sketch_prove} propose a ``draft–sketch–prove'' framework that first sketches proof outlines from informal arguments before completing subgoals with an automated prover. Collectively, these works highlight a growing toolkit of data generation, model training, and feedback mechanisms aimed at closing the gap in autoformalisation. However, these work fail to improve models using test-time compute which we tackle with \leanfire.

Conjecturing in the broader sense has been aimed to formalise open-ended conjectures to encourage mathematical discovery \citep{chau2025algebraic_combi}. Methodologically, many approaches interleave conjecturing with proving, where a placeholder conjecture is proposed and subsequently validated by a prover \citep{dong2025stp}. \citet{sun_etal_2025_enumerate-conjecture-prove} extend this idea by iteratively generating special coded cases from an autoformalised statement, forming candidate conjectures that are then tested by a prover in a repeated cycle. \citet{Zhou2024dontTrustVerify} demonstrate that for simple enough problems, LLMs could be used to generate the solutions and autoformalisation can verify them. However, we are the first to explicitly extract and evaluate conjecturing.

\section{Conclusion}
In this work, we identify conjecturing as an overlooked step in formal mathematical reasoning with LLMs, challenging the prevailing assumption that autoformalisation is a straightforward translation task. By introducing \ConjectureBench, a benchmark specifically designed to evaluate conjecture generation, and by proposing new metrics that disentangle conjecturing from autoformalisation, we provide the first systematic framework to measure and analyse this capability. Our results show that existing models substantially underperform when conjectures are withheld, revealing that much of their perceived success depends on having solutions pre-specified.
To address this gap, we develop \leanfire, an inference-time strategy that integrates informal Chain-of-Thought with formal Lean-of-Thought reasoning. This method enables the first successful end-to-end autoformalisation of PutnamBench ``no-answer'' problems, demonstrating that LLMs possess latent mathematical knowledge but require structured guidance to effectively conjecture and formalise. Manual analysis also identify two challenges: data contamination of existing benchmarks, and the task of generating useful definitions, functions and lemmata that would help autoformalisation, conjecturing and proving.
For future work, we argue that progress in formal mathematical reasoning hinges on treating conjecturing as an independent task. This calls for the development of richer conjecturing datasets, improved inference-time techniques, and training strategies that explicitly separate and then reintegrate conjecturing with autoformalisation.

\bibliography{iclr2026_conference}

\begin{thebibliography}{31}
\providecommand{\natexlab}[1]{#1}
\providecommand{\url}[1]{\texttt{#1}}
\expandafter\ifx\csname urlstyle\endcsname\relax
  \providecommand{\doi}[1]{doi: #1}\else
  \providecommand{\doi}{doi: \begingroup \urlstyle{rm}\Url}\fi

\bibitem[Achiam et~al.(2023)Achiam, Adler, Agarwal, Ahmad, Akkaya, Aleman, Almeida, Altenschmidt, Altman, Anadkat, et~al.]{achiam2023gpt}
Josh Achiam, Steven Adler, Sandhini Agarwal, Lama Ahmad, Ilge Akkaya, Florencia~Leoni Aleman, Diogo Almeida, Janko Altenschmidt, Sam Altman, Shyamal Anadkat, et~al.
\newblock {GPT-4 Technical Report}.
\newblock \emph{arXiv preprint arXiv:2303.08774}, 2023.
\newblock URL \url{https://arxiv.org/pdf/2303.08774}.

\bibitem[{AlphaProof and AlphaGeometry teams}(2024)]{AlphaProof2024}
{AlphaProof and AlphaGeometry teams}.
\newblock {AI achieves silver-medal standard solving International Mathematical Olympiad problems}.
\newblock \emph{Google DeepMind}, Jul 2024.
\newblock URL \url{https://deepmind.google/discover/blog/ai-solves-imo-problems-at-silver-medal-level/}.

\bibitem[Azerbayev et~al.(2023)Azerbayev, Piotrowski, Schoelkopf, Ayers, Radev, and Avigad]{azerbayev2023proofnet}
Zhangir Azerbayev, Bartosz Piotrowski, Hailey Schoelkopf, Edward~W. Ayers, Dragomir Radev, and Jeremy Avigad.
\newblock {ProofNet: Autoformalizing and Formally Proving Undergraduate-Level Mathematics}.
\newblock 2023.
\newblock URL \url{https://arxiv.org/abs/2302.12433}.

\bibitem[Cai et~al.(2024)Cai, Cao, Chen, Chen, Chen, Chen, Chen, Chen, Chen, Chu, Dong, Duan, Fan, Fei, Gao, Ge, Gu, Gu, Gui, Guo, Guo, He, Hu, Huang, Jiang, Jiao, Jin, Lei, Li, Li, Li, Li, Li, Li, Liu, Liu, Hong, Liu, Liu, Liu, Lv, Lv, Lv, Ma, Ma, Ma, Ning, Ouyang, Qiu, Qu, Shang, Shao, Song, Song, Sui, Sun, Sun, Tang, Wang, Wang, Wang, Wang, Wang, Wang, Wang, Wei, Weng, Wu, Xiong, and et~al.]{internlm2}
Zheng Cai, Maosong Cao, Haojiong Chen, Kai Chen, Keyu Chen, Xin Chen, Xun Chen, Zehui Chen, Zhi Chen, Pei Chu, Xiaoyi Dong, Haodong Duan, Qi~Fan, Zhaoye Fei, Yang Gao, Jiaye Ge, Chenya Gu, Yuzhe Gu, Tao Gui, Aijia Guo, Qipeng Guo, Conghui He, Yingfan Hu, Ting Huang, Tao Jiang, Penglong Jiao, Zhenjiang Jin, Zhikai Lei, Jiaxing Li, Jingwen Li, Linyang Li, Shuaibin Li, Wei Li, Yining Li, Hongwei Liu, Jiangning Liu, Jiawei Hong, Kaiwen Liu, Kuikun Liu, Xiaoran Liu, Chengqi Lv, Haijun Lv, Kai Lv, Li~Ma, Runyuan Ma, Zerun Ma, Wenchang Ning, Linke Ouyang, Jiantao Qiu, Yuan Qu, Fukai Shang, Yunfan Shao, Demin Song, Zifan Song, Zhihao Sui, Peng Sun, Yu~Sun, Huanze Tang, Bin Wang, Guoteng Wang, Jiaqi Wang, Jiayu Wang, Rui Wang, Yudong Wang, Ziyi Wang, Xingjian Wei, Qizhen Weng, Fan Wu, Yingtong Xiong, and et~al.
\newblock {InternLM2 Technical Report}.
\newblock \emph{arXiv preprint arXiv:2403.17297}, 2024.
\newblock URL \url{https://arxiv.org/abs/2403.17297}.

\bibitem[Chau et~al.(2025)Chau, Jenne, Brown, He, Raugas, Billey, and Kvinge]{chau2025algebraic_combi}
Herman Chau, Helen Jenne, Davis Brown, Jesse He, Mark Raugas, Sara~C. Billey, and Henry Kvinge.
\newblock {Machine Learning meets Algebraic Combinatorics: A Suite of Datasets Capturing Research-level Conjecturing Ability in Pure Mathematics}.
\newblock In \emph{Forty-second International Conference on Machine Learning}, 2025.
\newblock URL \url{https://openreview.net/forum?id=tlniJJFUW2}.

\bibitem[Chen et~al.(2025)Chen, Gu, Huang, Huang, Jiang, Jie, Jin, Jin, Li, Ma, Ren, Shen, Shi, Sun, Sun, Wang, Wang, Wang, Wei, Wei, Wu, Wu, Xia, Xin, Yang, Ying, Yuan, Yuan, Zhan, Zhang, Zhang, Zhang, Zhao, Zhao, Zhou, and Zhu]{chen2025seedproverdeepbroadreasoning}
Luoxin Chen, Jinming Gu, Liankai Huang, Wenhao Huang, Zhicheng Jiang, Allan Jie, Xiaoran Jin, Xing Jin, Chenggang Li, Kaijing Ma, Cheng Ren, Jiawei Shen, Wenlei Shi, Tong Sun, He~Sun, Jiahui Wang, Siran Wang, Zhihong Wang, Chenrui Wei, Shufa Wei, Yonghui Wu, Yuchen Wu, Yihang Xia, Huajian Xin, Fan Yang, Huaiyuan Ying, Hongyi Yuan, Zheng Yuan, Tianyang Zhan, Chi Zhang, Yue Zhang, Ge~Zhang, Tianyun Zhao, Jianqiu Zhao, Yichi Zhou, and Thomas~Hanwen Zhu.
\newblock {Seed-Prover: Deep and Broad Reasoning for Automated Theorem Proving}.
\newblock 2025.
\newblock URL \url{https://arxiv.org/abs/2507.23726}.

\bibitem[DeepSeek-AI et~al.(2024)DeepSeek-AI, Liu, Feng, Xue, Wang, Wu, Lu, Zhao, Deng, Zhang, Ruan, Dai, Guo, Yang, Chen, Ji, Li, Lin, Dai, Luo, Hao, Chen, Li, Zhang, Bao, Xu, Wang, Zhang, Ding, Xin, Gao, Li, Qu, Cai, Liang, Guo, Ni, Li, Wang, Chen, Chen, Yuan, Qiu, Li, Song, Dong, Hu, Gao, Guan, Huang, Yu, Wang, Zhang, Xu, Xia, Zhao, Wang, Zhang, Li, Wang, Zhang, Zhang, Tang, Li, Tian, Huang, Wang, Zhang, Wang, Zhu, Chen, Du, Chen, Jin, Ge, Zhang, Pan, Wang, Xu, Zhang, Chen, Li, Lu, Zhou, Chen, Wu, Ye, Ye, Ma, Wang, Zhou, Yu, Zhou, Pan, Wang, Yun, Pei, Sun, Xiao, Zeng, Zhao, An, Liu, Liang, Gao, Yu, Zhang, Li, Jin, Wang, Bi, Liu, Wang, Shen, Chen, Zhang, Chen, Nie, Sun, Wang, Cheng, Liu, Xie, Liu, Yu, Song, Shan, Zhou, Yang, Li, Su, Lin, Li, Wang, Wei, Zhu, Zhang, Xu, Xu, Huang, Li, Zhao, Sun, Li, Wang, Yu, Zheng, Zhang, Shi, Xiong, He, Tang, Piao, Wang, Tan, Ma, Liu, Guo, Wu, Ou, Zhu, Wang, Gong, Zou, He, Zha, Xiong, Ma, Yan, Luo, You, Liu, Zhou, Wu, Ren, Ren, Sha, Fu, Xu, Huang, Zhang, Xie, Zhang, Hao, Gou, Ma, Yan, Shao, Xu, Wu, Zhang, Li, Gu, Zhu, Liu, Li, Xie, Song, Gao, and Pan]{liu2024deepseek}
DeepSeek-AI, Aixin Liu, Bei Feng, Bing Xue, Bingxuan Wang, Bochao Wu, Chengda Lu, Chenggang Zhao, Chengqi Deng, Chenyu Zhang, Chong Ruan, Damai Dai, Daya Guo, Dejian Yang, Deli Chen, Dongjie Ji, Erhang Li, Fangyun Lin, Fucong Dai, Fuli Luo, Guangbo Hao, Guanting Chen, Guowei Li, H.~Zhang, Han Bao, Hanwei Xu, Haocheng Wang, Haowei Zhang, Honghui Ding, Huajian Xin, Huazuo Gao, Hui Li, Hui Qu, J.~L. Cai, Jian Liang, Jianzhong Guo, Jiaqi Ni, Jiashi Li, Jiawei Wang, Jin Chen, Jingchang Chen, Jingyang Yuan, Junjie Qiu, Junlong Li, Junxiao Song, Kai Dong, Kai Hu, Kaige Gao, Kang Guan, Kexin Huang, Kuai Yu, Lean Wang, Lecong Zhang, Lei Xu, Leyi Xia, Liang Zhao, Litong Wang, Liyue Zhang, Meng Li, Miaojun Wang, Mingchuan Zhang, Minghua Zhang, Minghui Tang, Mingming Li, Ning Tian, Panpan Huang, Peiyi Wang, Peng Zhang, Qiancheng Wang, Qihao Zhu, Qinyu Chen, Qiushi Du, R.~J. Chen, R.~L. Jin, Ruiqi Ge, Ruisong Zhang, Ruizhe Pan, Runji Wang, Runxin Xu, Ruoyu Zhang, Ruyi Chen, S.~S. Li, Shanghao Lu, Shangyan Zhou, Shanhuang Chen, Shaoqing Wu, Shengfeng Ye, Shengfeng Ye, Shirong Ma, Shiyu Wang, Shuang Zhou, Shuiping Yu, Shunfeng Zhou, Shuting Pan, T.~Wang, Tao Yun, Tian Pei, Tianyu Sun, W.~L. Xiao, Wangding Zeng, Wanjia Zhao, Wei An, Wen Liu, Wenfeng Liang, Wenjun Gao, Wenqin Yu, Wentao Zhang, X.~Q. Li, Xiangyue Jin, Xianzu Wang, Xiao Bi, Xiaodong Liu, Xiaohan Wang, Xiaojin Shen, Xiaokang Chen, Xiaokang Zhang, Xiaosha Chen, Xiaotao Nie, Xiaowen Sun, Xiaoxiang Wang, Xin Cheng, Xin Liu, Xin Xie, Xingchao Liu, Xingkai Yu, Xinnan Song, Xinxia Shan, Xinyi Zhou, Xinyu Yang, Xinyuan Li, Xuecheng Su, Xuheng Lin, Y.~K. Li, Y.~Q. Wang, Y.~X. Wei, Y.~X. Zhu, Yang Zhang, Yanhong Xu, Yanhong Xu, Yanping Huang, Yao Li, Yao Zhao, Yaofeng Sun, Yaohui Li, Yaohui Wang, Yi~Yu, Yi~Zheng, Yichao Zhang, Yifan Shi, Yiliang Xiong, Ying He, Ying Tang, Yishi Piao, Yisong Wang, Yixuan Tan, Yiyang Ma, Yiyuan Liu, Yongqiang Guo, Yu~Wu, Yuan Ou, Yuchen Zhu, Yuduan Wang, Yue Gong, Yuheng Zou, Yujia He, Yukun Zha, Yunfan Xiong, Yunxian Ma, Yuting Yan, Yuxiang Luo, Yuxiang You, Yuxuan Liu, Yuyang Zhou, Z.~F. Wu, Z.~Z. Ren, Zehui Ren, Zhangli Sha, Zhe Fu, Zhean Xu, Zhen Huang, Zhen Zhang, Zhenda Xie, Zhengyan Zhang, Zhewen Hao, Zhibin Gou, Zhicheng Ma, Zhigang Yan, Zhihong Shao, Zhipeng Xu, Zhiyu Wu, Zhongyu Zhang, Zhuoshu Li, Zihui Gu, Zijia Zhu, Zijun Liu, Zilin Li, Ziwei Xie, Ziyang Song, Ziyi Gao, and Zizheng Pan.
\newblock {Deepseek-v3 Technical Report}.
\newblock \emph{arXiv preprint arXiv:2412.19437}, 2024.

\bibitem[Dong \& Ma(2025)Dong and Ma]{dong2025stp}
Kefan Dong and Tengyu Ma.
\newblock {STP}: Self-play {LLM} theorem provers with iterative conjecturing and proving.
\newblock In \emph{Forty-second International Conference on Machine Learning}, 2025.
\newblock URL \url{https://openreview.net/forum?id=zWArMedNuW}.

\bibitem[Fernández-León et~al.(2021)Fernández-León, Gavilán-Izquierdo, and Toscano]{Fernández-León28052021_proofVSconjecture}
Aurora Fernández-León, José~María Gavilán-Izquierdo, and Rocío Toscano.
\newblock A case study of the practices of conjecturing and proving of research mathematicians.
\newblock \emph{International Journal of Mathematical Education in Science and Technology}, 52\penalty0 (5):\penalty0 767--781, 2021.
\newblock \doi{10.1080/0020739X.2020.1717658}.
\newblock URL \url{https://doi.org/10.1080/0020739X.2020.1717658}.

\bibitem[Gouëzel \& Shchur(2019)Gouëzel and Shchur]{Gouezel-Shchur-2019-lean_error}
Sébastien Gouëzel and Vladimir Shchur.
\newblock {A corrected quantitative version of the Morse lemma}.
\newblock \emph{Journal of Functional Analysis}, 277\penalty0 (4):\penalty0 1258--1268, 2019.
\newblock ISSN 0022-1236.
\newblock \doi{https://doi.org/10.1016/j.jfa.2019.02.021}.
\newblock URL \url{https://www.sciencedirect.com/science/article/pii/S0022123619300801}.

\bibitem[Jiang et~al.(2023)Jiang, Welleck, Zhou, Lacroix, Liu, Li, Jamnik, Lample, and Wu]{jiang2023draft_sketch_prove}
Albert~Qiaochu Jiang, Sean Welleck, Jin~Peng Zhou, Timothee Lacroix, Jiacheng Liu, Wenda Li, Mateja Jamnik, Guillaume Lample, and Yuhuai Wu.
\newblock {Draft, Sketch, and Prove: Guiding Formal Theorem Provers with Informal Proofs}.
\newblock In \emph{The Eleventh International Conference on Learning Representations}, 2023.
\newblock URL \url{https://openreview.net/forum?id=SMa9EAovKMC}.

\bibitem[Lin et~al.(2025)Lin, Tang, Lyu, Wu, Lin, Yang, LI, Xia, Chen, Arora, and Jin]{lin2025goedel1}
Yong Lin, Shange Tang, Bohan Lyu, Jiayun Wu, Hongzhou Lin, Kaiyu Yang, Jia LI, Mengzhou Xia, Danqi Chen, Sanjeev Arora, and Chi Jin.
\newblock {Goedel-Prover: A Frontier Model for Open-Source Automated Theorem Proving}.
\newblock In \emph{Second Conference on Language Modeling}, 2025.
\newblock URL \url{https://openreview.net/forum?id=x2y9i2HDjD}.

\bibitem[Liu et~al.(2025{\natexlab{a}})Liu, Lin, Bayer, Dillies, Jiang, Liang, Soletskyi, Wang, Xie, Xiong, et~al.]{liu2505combibench}
J~Liu, X~Lin, J~Bayer, Y~Dillies, W~Jiang, X~Liang, R~Soletskyi, H~Wang, Y~Xie, B~Xiong, et~al.
\newblock {Combibench: Benchmarking llm capability for combinatorial mathematics}.
\newblock \emph{arXiv preprint arXiv:2505.03171}, 2025{\natexlab{a}}.
\newblock URL \url{https://arxiv.org/pdf/2505.03171}.

\bibitem[Liu et~al.(2025{\natexlab{b}})Liu, Zheng, Lu, Cao, and Yan]{liu2025rethinking}
Qi~Liu, Xinhao Zheng, Xudong Lu, Qinxiang Cao, and Junchi Yan.
\newblock {Rethinking and Improving Autoformalization: Towards a Faithful Metric and a Dependency Retrieval-based Approach}.
\newblock In \emph{The Thirteenth International Conference on Learning Representations}, 2025{\natexlab{b}}.
\newblock URL \url{https://openreview.net/forum?id=hUb2At2DsQ}.

\bibitem[Lu et~al.(2024)Lu, Wan, Liu, Huang, Xiong, Liu, Shen, Jin, Zhang, Wang, Yang, Tang, and Guo]{lu2024processdrivenautoformalizationlean4}
Jianqiao Lu, Yingjia Wan, Zhengying Liu, Yinya Huang, Jing Xiong, Chengwu Liu, Jianhao Shen, Hui Jin, Jipeng Zhang, Haiming Wang, Zhicheng Yang, Jing Tang, and Zhijiang Guo.
\newblock {Process-Driven Autoformalization in Lean 4}, 2024.
\newblock URL \url{https://arxiv.org/abs/2406.01940}.

\bibitem[Metz(2025)]{Gemini-2025-IMOs}
Cade Metz.
\newblock {Google A.I. System Wins Gold Medal in International Math Olympiad}.
\newblock \emph{The New York Times}, Jul 2025.
\newblock URL \url{https://www.nytimes.com/2025/07/21/technology/google-ai-international-mathematics-olympiad.html}.

\bibitem[Moura \& Ullrich(2021)Moura and Ullrich]{moura2021lean}
Leonardo~de Moura and Sebastian Ullrich.
\newblock {The Lean 4 theorem prover and programming language}.
\newblock In \emph{Automated Deduction--CADE 28: 28th International Conference on Automated Deduction, Virtual Event, July 12--15, 2021, Proceedings 28}, pp.\  625--635. Springer, 2021.

\bibitem[Pauli(2022)]{math_defintions}
Sebastian Pauli.
\newblock Mat 112 integers and modern applications for the uninitiated.
\newblock \emph{Caesar Ciphers}, 2022.
\newblock URL \url{https://mat112.uncg.edu/HTML/root-1-2.html}.

\bibitem[Peng et~al.(2025)Peng, Yao, Ma, Guo, Li, Zhang, Zhang, Zhang, Yu, Li, et~al.]{peng2025criticlean}
Zhongyuan Peng, Yifan Yao, Kaijing Ma, Shuyue Guo, Yizhe Li, Yichi Zhang, Chenchen Zhang, Yifan Zhang, Zhouliang Yu, Luming Li, et~al.
\newblock {Criticlean: Critic-guided reinforcement learning for mathematical formalization}.
\newblock \emph{arXiv preprint arXiv:2507.06181}, 2025.
\newblock URL \url{https://arxiv.org/pdf/2507.06181}.

\bibitem[Poiroux et~al.(2025)Poiroux, Weiss, Kunčak, and Bosselut]{poiroux2025improvingautoformalizationusingtype}
Auguste Poiroux, Gail Weiss, Viktor Kunčak, and Antoine Bosselut.
\newblock {Improving Autoformalization using Type Checking}, 2025.
\newblock URL \url{https://arxiv.org/abs/2406.07222}.

\bibitem[Sun et~al.(2025)Sun, Tang, Li, Maddison, and Meel]{sun_etal_2025_enumerate-conjecture-prove}
Jialiang Sun, Yuzhi Tang, Ao~Li, Chris~J Maddison, and Kuldeep~S Meel.
\newblock {Enumerate-Conjecture-Prove: Formally Solving Answer-Construction Problems in Math Competitions}.
\newblock \emph{arXiv preprint arXiv:2505.18492}, 2025.
\newblock URL \url{https://arxiv.org/abs/2505.18492}.

\bibitem[Szegedy(2020)]{szegedy2020position}
Christian Szegedy.
\newblock {A Promising Path Towards Autoformalization and General Artificial Intelligence}.
\newblock In \emph{Intelligent Computer Mathematics: 13th International Conference, CICM 2020, Bertinoro, Italy, July 26–31, 2020, Proceedings}, pp.\  3–20, Berlin, Heidelberg, 2020. Springer-Verlag.
\newblock ISBN 978-3-030-53517-9.
\newblock \doi{10.1007/978-3-030-53518-6_1}.
\newblock URL \url{https://doi.org/10.1007/978-3-030-53518-6_1}.

\bibitem[Tsoukalas et~al.(2024)Tsoukalas, Lee, Jennings, Xin, Ding, Jennings, Thakur, and Chaudhuri]{putnam_tsoukalas_etal_2024}
George Tsoukalas, Jasper Lee, John Jennings, Jimmy Xin, Michelle Ding, Michael Jennings, Amitayush Thakur, and Swarat Chaudhuri.
\newblock {PutnamBench: Evaluating Neural Theorem-Provers on the Putnam Mathematical Competition}.
\newblock In A.~Globerson, L.~Mackey, D.~Belgrave, A.~Fan, U.~Paquet, J.~Tomczak, and C.~Zhang (eds.), \emph{Advances in Neural Information Processing Systems}, volume~37, pp.\  11545--11569. Curran Associates, Inc., 2024.
\newblock URL \url{https://proceedings.neurips.cc/paper_files/paper/2024/file/1582eaf9e0cf349e1e5a6ee453100aa1-Paper-Datasets_and_Benchmarks_Track.pdf}.

\bibitem[Wang et~al.(2025)Wang, Unsal, Lin, Baksys, Liu, Santos, Sung, Vinyes, Ying, Zhu, Lu, de~Saxcé, Bailey, Song, Xiao, Zhang, Zhang, Pu, Zhu, Liu, Bayer, Michel, Yu, Dreyfus-Schmidt, Tunstall, Pagani, Machado, Bourigault, Wang, Polu, Barroyer, Li, Niu, Fleureau, Hu, Yu, Wang, Yang, Liu, and Li]{wang2025kimina}
Haiming Wang, Mert Unsal, Xiaohan Lin, Mantas Baksys, Junqi Liu, Marco~Dos Santos, Flood Sung, Marina Vinyes, Zhenzhe Ying, Zekai Zhu, Jianqiao Lu, Hugues de~Saxcé, Bolton Bailey, Chendong Song, Chenjun Xiao, Dehao Zhang, Ebony Zhang, Frederick Pu, Han Zhu, Jiawei Liu, Jonas Bayer, Julien Michel, Longhui Yu, Léo Dreyfus-Schmidt, Lewis Tunstall, Luigi Pagani, Moreira Machado, Pauline Bourigault, Ran Wang, Stanislas Polu, Thibaut Barroyer, Wen-Ding Li, Yazhe Niu, Yann Fleureau, Yangyang Hu, Zhouliang Yu, Zihan Wang, Zhilin Yang, Zhengying Liu, and Jia Li.
\newblock Kimina-prover preview: Towards large formal reasoning models with reinforcement learning.
\newblock \emph{arXiv preprint arXiv:2504.11354}, 2025.
\newblock URL \url{https://arxiv.org/abs/2504.11354}.

\bibitem[Wenzel et~al.(2008)Wenzel, Paulson, and Nipkow]{wenzel_etal_2008_isabelle}
Makarius Wenzel, Lawrence~C. Paulson, and Tobias Nipkow.
\newblock The isabelle framework.
\newblock In Otmane~Ait Mohamed, C{\'e}sar Mu{\~{n}}oz, and Sofi{\`e}ne Tahar (eds.), \emph{Theorem Proving in Higher Order Logics}, pp.\  33--38, Berlin, Heidelberg, 2008. Springer Berlin Heidelberg.
\newblock ISBN 978-3-540-71067-7.

\bibitem[Xin et~al.(2024)Xin, Guo, Shao, Ren, Zhu, Liu, Ruan, Li, and Liang]{xin2024deepseekprover}
Huajian Xin, Daya Guo, Zhihong Shao, Z.Z. Ren, Qihao Zhu, Bo~Liu, Chong Ruan, Wenda Li, and Xiaodan Liang.
\newblock Advancing theorem proving in {LLM}s through large-scale synthetic data.
\newblock In \emph{The 4th Workshop on Mathematical Reasoning and AI at NeurIPS'24}, 2024.
\newblock URL \url{https://openreview.net/forum?id=TPtXLihkny}.

\bibitem[Yang et~al.(2025{\natexlab{a}})Yang, Li, Yang, Zhang, Hui, Zheng, Yu, Gao, Huang, Lv, Zheng, Liu, Zhou, Huang, Hu, Ge, Wei, Lin, Tang, Yang, Tu, Zhang, Yang, Yang, Zhou, Zhou, Lin, Dang, Bao, Yang, Yu, Deng, Li, Xue, Li, Zhang, Wang, Zhu, Men, Gao, Liu, Luo, Li, Tang, Yin, Ren, Wang, Zhang, Ren, Fan, Su, Zhang, Zhang, Wan, Liu, Wang, Cui, Zhang, Zhou, and Qiu]{yang2025qwen3technicalreport}
An~Yang, Anfeng Li, Baosong Yang, Beichen Zhang, Binyuan Hui, Bo~Zheng, Bowen Yu, Chang Gao, Chengen Huang, Chenxu Lv, Chujie Zheng, Dayiheng Liu, Fan Zhou, Fei Huang, Feng Hu, Hao Ge, Haoran Wei, Huan Lin, Jialong Tang, Jian Yang, Jianhong Tu, Jianwei Zhang, Jianxin Yang, Jiaxi Yang, Jing Zhou, Jingren Zhou, Junyang Lin, Kai Dang, Keqin Bao, Kexin Yang, Le~Yu, Lianghao Deng, Mei Li, Mingfeng Xue, Mingze Li, Pei Zhang, Peng Wang, Qin Zhu, Rui Men, Ruize Gao, Shixuan Liu, Shuang Luo, Tianhao Li, Tianyi Tang, Wenbiao Yin, Xingzhang Ren, Xinyu Wang, Xinyu Zhang, Xuancheng Ren, Yang Fan, Yang Su, Yichang Zhang, Yinger Zhang, Yu~Wan, Yuqiong Liu, Zekun Wang, Zeyu Cui, Zhenru Zhang, Zhipeng Zhou, and Zihan Qiu.
\newblock {Qwen3 Technical Report}.
\newblock 2025{\natexlab{a}}.
\newblock URL \url{https://arxiv.org/pdf/2505.09388}.

\bibitem[Yang et~al.(2025{\natexlab{b}})Yang, Poesia, He, Li, Lauter, Chaudhuri, and Song]{yang2025position_formal_mathematical_reasoning}
Kaiyu Yang, Gabriel Poesia, Jingxuan He, Wenda Li, Kristin~E. Lauter, Swarat Chaudhuri, and Dawn Song.
\newblock Position: Formal mathematical reasoning{\textemdash}a new frontier in {AI}.
\newblock In \emph{Forty-second International Conference on Machine Learning Position Paper Track}, 2025{\natexlab{b}}.
\newblock URL \url{https://openreview.net/forum?id=HuvAM5x2xG}.

\bibitem[Ying et~al.(2024)Ying, Wu, Geng, Wang, Lin, and Chen]{ying2024lean_workbook}
Huaiyuan Ying, Zijian Wu, Yihan Geng, JIayu Wang, Dahua Lin, and Kai Chen.
\newblock {Lean Workbook: A large-scale Lean problem set formalized from natural language math problems}.
\newblock In \emph{The Thirty-eight Conference on Neural Information Processing Systems Datasets and Benchmarks Track}, 2024.
\newblock URL \url{https://openreview.net/forum?id=Vcw3vzjHDb}.

\bibitem[Zheng et~al.(2023)Zheng, Chiang, Sheng, Zhuang, Wu, Zhuang, Lin, Li, Li, Xing, Zhang, Gonzalez, and Stoica]{zheng2023llm-as-a-judge}
Lianmin Zheng, Wei-Lin Chiang, Ying Sheng, Siyuan Zhuang, Zhanghao Wu, Yonghao Zhuang, Zi~Lin, Zhuohan Li, Dacheng Li, Eric Xing, Hao Zhang, Joseph~E Gonzalez, and Ion Stoica.
\newblock {Judging LLM-as-a-Judge with MT-Bench and Chatbot Arena}.
\newblock In A.~Oh, T.~Naumann, A.~Globerson, K.~Saenko, M.~Hardt, and S.~Levine (eds.), \emph{Advances in Neural Information Processing Systems}, volume~36, pp.\  46595--46623. Curran Associates, Inc., 2023.
\newblock URL \url{https://proceedings.neurips.cc/paper_files/paper/2023/file/91f18a1287b398d378ef22505bf41832-Paper-Datasets_and_Benchmarks.pdf}.

\bibitem[Zhou et~al.(2024)Zhou, Staats, Li, Szegedy, Weinberger, and Wu]{Zhou2024dontTrustVerify}
Jin~Peng Zhou, Charles Staats, Wenda Li, Christian Szegedy, Kilian~Q. Weinberger, and Yuhuai Wu.
\newblock {Don't Trust: Verify - Grounding LLM Quantitative Reasoning with Autoformalization}.
\newblock In \emph{ICLR}, 2024.
\newblock URL \url{https://openreview.net/forum?id=V5tdi14ple}.

\end{thebibliography}
\bibliographystyle{iclr2026_conference}

\newpage
\appendix

\section{\leanfire}
In this Appendix section, we provide the five examples used both as seed questions and few-shot examples \ref{app: seed-questions}. We also include the prompts used to generate the CoT and the subsequent LoT in \ref{app:cot-lot-instructions}.

\subsection{Seed Questions}\label{app: seed-questions}
\lstdefinestyle{leanfirelisting}{
  basicstyle=\ttfamily\small\color{black},
  keywordstyle=\color{black},
  stringstyle=\color{black},
  commentstyle=\color{black},
  identifierstyle=\color{black},
  morekeywords={},
  literate=
    {=}{{=}}1
    {<}{{<}}1
    {>}{{>}}1
    {∧}{{\wedge}}1
    {≤}{{\le}}1
    {∑}{{\sum}}1,
  mathescape=true,
  columns=fullflexible,
  keepspaces=true,
  breaklines=true,
  breakindent=11pt,       
  xleftmargin=0pt,       
  showstringspaces=false
}
\begin{figure}[h!]
\centering
\footnotesize
\begin{tabular}{p{13cm}}
\paragraph{Seed/Few-shot example 1 of 5}
\\
\toprule
\textbf{Name} \\
putnam\_2004\_a1\\
\midrule
\textbf{Informal Statement} \\
Basketball star Shanille O'Keal's team statistician keeps track of the number, $S(N)$, of successful free throws she has made in her first $N$ attempts of the season. Early in the season, $S(N)$ was less than $80\%$ of $N$, but by the end of the season, $S(N)$ was more than $80\%$ of $N$. Proof or disprove that it there necessarily was a moment in between when $S(N)$ was exactly $80\%$ of $N$. \\

\midrule

\textbf{LeanFIRe Reasoning}
\begin{lstlisting}[style=leanfirelisting]
- Each attempt has a value in $\{0,1\}$, 0 for fail, 1 for success, i.e. attempt: $\mathbb{N} \rightarrow \{0,1\}$.
  Lean: attempt : $\mathbb{N}$ → Fin 2

- The function S is the average score of the attempt, i.e., the sum of the attempts divided by the number of attempts S: attempts → $\mathbb{R}$.
  Lean: S : ($\mathbb{N}$ → Fin 2) → $\mathbb{N}$ → $\mathbb{R}$
    S attempts N = ($\sum$ {i : Fin N} (attempts i).1) / N

- S(N) can be written as S(N) = m_N / N where m_N is the number of successes in N tries, i.e. m_N = $\sum_{i=1}^N \1_{success}$.
  Lean: (encoded in definition of S above)

- The success rate is below 80% at one point a, S(a) < 0.8, and above 80% at another point b > a, S(b) > 0.8.
  Lean: 1 $\le$ a $\wedge$ a < b $\wedge$ S attempts a < 0.8 $\wedge$ S attempts b > 0.8

- Show there exists c $\in$ (a,b) with S(c) = 0.8.
  Lean: $\exists$ c : $\mathbb{N}$, a < c $\wedge$ c < b $\wedge$ S attempts c = 0.8
\end{lstlisting}
\\
\midrule

\textbf{Conjecture}
\begin{lstlisting}
abbrev conjecture : Prop := True
\end{lstlisting}
\\
\midrule

\textbf{Formal Statement}
\begin{lstlisting}
theorem putnam_2004_a1
  (S : (ℕ → Fin 2) → ℕ → ℝ)
  (hS : ∀ attempts, ∀ N ≥ 1, S attempts N = (∑ i : Fin N, (attempts i).1) / N) :
  (∀ attempts a b,
    (1 ≤ a ∧ a < b ∧ S attempts a < 0.8 ∧ S attempts b > 0.8) →
      (∃ c : ℕ, a < c ∧ c < b ∧ S attempts c = 0.8))
  ↔ conjecture :=
sorry
\end{lstlisting}
 \\
\bottomrule

\end{tabular}
\caption{Example (1/5) from Putnam annotated with informal and formal hint reasoning steps.}
    \label{fig:seed-0}
\end{figure}

\begin{figure}[t!]
\centering
\footnotesize
\begin{tabular}{p{13cm}}
\paragraph{Seed/Few-shot example 2 of 5}
\\
\toprule
\textbf{Name} \\
putnam\_2009\_b2\\
\midrule
\textbf{Informal Statement} \\
A game involves jumping to the right on the real number line. If $a$ and $b$ are real numbers and $b > a$, the cost of jumping from $a$ to $b$ is $b^3-ab^2$. For what real numbers $c$ can one travel from $0$ to $1$ in a finite number of jumps with total cost exactly $c$?
\\
\midrule

\textbf{LeanFIRe Reasoning}
\begin{lstlisting}[style=leanfirelisting]
- The jumps can be modelled as a sequence that partitions the interval (0,1),  with $N \in \mathbb{N}$ jumps, $s_0=0$, $s_i=1$, and $s_i < s_{i+1}$ for all $0 \le i < N$.
  Lean: s : Fin (N + 1) → $\mathbb{R}$
    validPath (s : Fin (N + 1) → $\mathbb{R}$) : Prop :=
    s 0 = 0 $\wedge$ s (Fin.last N) = 1 $\wedge$ $\forall$ i : Fin N, s i < s (i.succ)

- The cost of a jump from $s_i$ to $s_{i+1}$ is $s_{i+1}^3 - s_i * s_{i+1}^2$.
  Lean: jumpCost (a b : $\mathbb{R}$) : $\mathbb{R}$ := b^3 - a * b^2

- The total cost for all jumps is $\sum_{i=0}^{N-1} (s_{i+1}^3 - s_i * s_{i+1}^2)$.
  Lean: totalCost (s : Fin (N + 1) → $\mathbb{R}$) : $\mathbb{R}$ :=
    $\sum$ {i : Fin N} jumpCost (s i) (s (i.succ))

- The set of reachable costs is { c $\in$ $\mathbb{R}$ $\mid$ $\exists$ N $\in$ $\mathbb{N}$, validPath s $\wedge$ totalCost(s) = c }.
  Lean: reachableCosts : Set $\mathbb{R}$ :=
    {c : $\mathbb{R}$ $\mid$ $\exists$ (N : $\mathbb{N}$) (s : Fin (N + 1) → $\mathbb{R}$),
    validPath s $\wedge$ totalCost s = c}
\end{lstlisting}
\\
\midrule

\textbf{Conjecture}
\begin{lstlisting}
abbrev conjecture : Set ℝ := Ioc (1 / 3) 1
\end{lstlisting} 
\\
\midrule

\textbf{Formal Statement}
\begin{lstlisting}
theorem putnam_2009_b2
: ({c : ℝ | ∃ s : ℕ → ℝ, s 0 = 0 ∧ StrictMono s ∧ (∃ n : ℕ, s n = 1 ∧ ((∑ i ∈ Finset.range n, ((s (i + 1)) ^ 3 - (s i) * (s (i + 1)) ^ 2)) = c))} = conjecture) :=
sorry
\end{lstlisting}
\\
\bottomrule
\end{tabular}
\caption{Example (2/5) from Putnam annotated with informal and formal hint reasoning steps.}
    \label{fig:seed-1}
\vspace{8cm}
\end{figure}
\clearpage

\begin{figure}[h!]
\centering
\footnotesize
\begin{tabular}{p{13cm}}
\paragraph{Seed/Few-shot example 3 of 5}
\\
\toprule
\textbf{Name} \\
putnam\_2013\_b2\\
\midrule
\textbf{Informal Statement} \\
Let $C =  \bigcup_{N=1}^ \infty C_N$, where $C_N$ denotes the set of those `cosine polynomials' of the form \[ f(x) = 1 +  \sum_{n=1}^N a_n  \cos(2  \pi n x) \] for which: \begin{enumerate} \item[(i)] $f(x)  \geq 0$ for all real $x$, and \item[(ii)] $a_n = 0$ whenever $n$ is a multiple of $3$. \end{enumerate} Determine the maximum value of $f(0)$ as $f$ ranges through $C$, and prove that this maximum is attained.
\\
\midrule

\textbf{LeanFIRe Reasoning}
\begin{lstlisting}[style=leanfirelisting]
- C is the set of all C_N for a given $N \in \mathbb{N}$.
  Lean: C_N (N : $\mathbb{N}$) : Set ($\mathbb{R} \to \mathbb{R}$) :=
    { f | $\exists$ (a : $\mathbb{N} \to \mathbb{R}$),
    ($\forall$ x, f x = 1 + $\sum${n $\in$ Finset.range N} a n * Real.cos (2 * $\pi$ *
     n * x)) $\wedge$
    ($\forall$ x, f x $\ge$ 0) $\wedge$ ($\forall$ n, n % 3 = 0 → a n = 0) }

- C_N is defined as the set of polynomials of the form f(x) = 1 + $\sum_{n=1}^N a_n \cos(2\pi n x)$ where f(x) $\ge$ 0 for all x $\in \mathbb{R}$, and the coefficient $a_n=0$ whenever n is a multiple of 3.
  Lean: (above definition of C_N already encodes this)

- Therefore, C_N = { f(x) $\in \mathbb{R} \mid$ f(x) = 1 + $\sum_{n=1}^N a_n \cos(2\pi n x)$, f(x) $\ge$ 0  $\forall$ x $\in \mathbb{R}$, $a_n = 0$ if n $\bmod$ 3 = 0 }.
  Lean: (same C_N definition)

- C is the union of all the C_N, i.e. $C = \bigcup_{N=1}^{\infty} C_N$.
  Lean: C : Set ($\mathbb{R} \to \mathbb{R}$) := $\bigcup$ N, C_N N

- Determine the maximum f(0) within all possible C_N, i.e. $\sup$ { f(0) $\mid$ f $\in$ C }.
  Lean: supF0 : $\mathbb{R}$ := Sup { f 0 | f $\in$ C }
\end{lstlisting}
\\
\midrule

\textbf{Conjecture}
\begin{lstlisting}
abbrev conjecture : ℝ := 3
\end{lstlisting} 
\\
\midrule

\textbf{Formal Statement}
\begin{lstlisting}
theorem putnam_2013_b2
  (CN : ℕ → Set (ℝ → ℝ))
  (hCN : ∀ N : ℕ, CN N =
    {f : ℝ → ℝ |
      (∀ x : ℝ, f x ≥ 0) ∧
      ∃ a : List ℝ, a.length = N + 1 ∧ (∀ n : Fin (N + 1), 3 | (n : ℕ) → a[n]! = 0) ∧
      ∀ x : ℝ, f x = 1 + ∑ n ∈ Finset.Icc 1 N, a[(n : ℕ)]! * Real.cos (2*Real.pi*n*x)}) :
  IsGreatest {f 0 | f ∈ ∪ N ∈ Ici 1, CN N} conjecture :=
sorry
\end{lstlisting}
\\
\bottomrule

\end{tabular}
\caption{Example (3/5) from Putnam annotated with informal and formal hint reasoning steps.}
    \label{fig:seed-2}
\end{figure}

\begin{figure}[h!]
\centering
\footnotesize
\begin{tabular}{p{13cm}}
\paragraph{Seed/Few-shot example 4 of 5}
\\
\toprule
\textbf{Name} \\
putnam\_2014\_a2\\
\midrule
\textbf{Informal Statement} \\
Let $A$ be the $n  \times n$ matrix whose entry in the $i$-th row and $j$-th column is $ \frac{1}{ \min(i,j)}$ for $1  \leq i,j  \leq n$. Compute $ \det(A)$.
\\
\midrule

\textbf{LeanFIRe Reasoning}
\begin{lstlisting}[style=leanfirelisting]
- Let the dimension of the matrix be $n \in \mathbb{N}$, and the $n \times n$ matrix $A \in \mathbb{R}^{n \times n}$.
  Lean: A (n : $\mathbb{N}$) : Matrix (Fin n) (Fin n) $\mathbb{R}$ :=

- Define $A_{ij}$ to be the entry from the i-th row and j-th column of matrix A.
  Lean: (implicit in the matrix function arguments $\lambda$ i j)

- Set each entry to be the minimum between its column and row value, i.e. $A_{ij}$ = 1 / $\min(i, j)$ $\forall$ 1 $\le$ i, j $\le$ n.
  Lean: $\lambda$ i j => 1 / min (i.1 + 1) (j.1 + 1)
  Note: i.1 + 1 and j.1 + 1 are used because Lean indices start at 0 but min(i,j) starts at 1

- Evaluate $\det(A)$.
  Lean: detA (n : $\mathbb{N}$) : $\mathbb{R}$ := Matrix.det (A n)
\end{lstlisting}
\\
\midrule

\textbf{Conjecture}
\begin{lstlisting}
abbrev conjecture : ℝ := 3
\end{lstlisting} 
\\
\midrule

\textbf{Formal Statement}
\begin{lstlisting}
theorem putnam_2013_b2
  (CN : ℕ → Set (ℝ → ℝ))
  (hCN : ∀ N : ℕ, CN N =
    {f : ℝ → ℝ |
      (∀ x : ℝ, f x ≥ 0) ∧
      ∃ a : List ℝ, a.length = N + 1 ∧ (∀ n : Fin (N + 1), 3 | (n : ℕ) → a[n]! = 0) ∧
      ∀ x : ℝ, f x = 1 + ∑ n ∈ Finset.Icc 1 N, a[(n : ℕ)]! * Real.cos (2*Real.pi*n*x)}) :
  IsGreatest {f 0 | f ∈ ∪ N ∈ Ici 1, CN N} conjecture :=
sorry
\end{lstlisting}
\\
\bottomrule

\end{tabular}
\caption{Example (4/5) from Putnam annotated with informal and formal hint reasoning steps.}
    \label{fig:seed-3}
\vspace{8cm}
\end{figure}

\begin{figure}[h!]
\centering
\footnotesize
\begin{tabular}{p{13cm}}
\paragraph{Seed/Few-shot example 5 of 5}
\\
\toprule
\textbf{Name} \\
putnam\_2015\_a2\\
\midrule
\textbf{Informal Statement} \\
Let $a_0=1$, $a_1=2$, and $a_n=4a_{n-1}-a_{n-2}$ for $n  \geq 2$. Find an odd prime factor of $a_{2015}$.
\\
\midrule

\textbf{LeanFIRe Reasoning}
\begin{lstlisting}[style=leanfirelisting]
- A recurrence relation is initialised with 1 and 2 as the starting points, i.e. $a_0 = 1$ and $a_1 = 2$.
  Lean: a : $\mathbb{N} \to \mathbb{N}$
    a 0 = 1
    a 1 = 2

- It is defined as 4 times the previous term minus the term before the previous one, i.e. $a_n = 4 a_{n-1} - a_{n-2}$ for $n \ge 2$.
  Lean: $\forall n \ge 2, a\ n = 4 * a\ (n - 1) - a\ (n - 2)$

- For the 2015th term of the sequence, $a_{2015}$, determine a factor $c \in \mathbb{N}$such that:
  • $c \mid a_{2015}$
  • $c$ is odd ($\exists$ n $\in \mathbb{N}$, c = $2n - 1$)
  • $c$ is prime (no divisor $k > 1$ except itself)
  Lean: $\exists$ p : $\mathbb{N}$, p $\mid$ a 2015 $\wedge$ Nat.Prime p $\wedge$ Odd p
\end{lstlisting}
\\
\midrule

\textbf{Conjecture}
\begin{lstlisting}
abbrev conjecture : ℕ := 181
\end{lstlisting} 
\\
\midrule

\textbf{Formal Statement}
\begin{lstlisting}
theorem putnam_2015_a2
(a : ℕ → ℤ)
(abase : a 0 = 1 ∧ a 1 = 2)
(arec : ∀ n ≥ 2, a n = 4 * a (n - 1) - a (n - 2))
: Odd conjecture ∧ conjecture.Prime ∧ ((conjecture : ℤ) | a 2015) :=
sorry
\end{lstlisting}
\\
\bottomrule

\end{tabular}
\caption{Example (5/5) from Putnam annotated with informal and formal hint reasoning steps.}
    \label{fig:seed-4}
\vspace{10cm}
\end{figure}

\clearpage

\subsection{\leanfire Prompts} \label{app:cot-lot-instructions}
\paragraph{Chain-of-Thought (CoT) generation prompt}
\begin{figure}[h!]
\centering
\footnotesize
\begin{tabular}{p{13cm}}
\paragraph{Chain-of-Thought (CoT) Generation Prompt}
\\
\toprule
\textbf{System Prompt} \\
\begin{verbnobox}[\fontsize{8pt}{8pt}\selectfont]
You are an advanced assistant specializing in formal mathematics and Lean 4 
theorem proving. You have extensive expertise in translating mathematical 
concepts from natural language into precise Lean 4 code.

\end{verbnobox}

 \\ \midrule

\textbf{User Prompt} \\

\begin{verbnobox}[\fontsize{8pt}{8pt}\selectfont]
Using the provided informal statement, write a concise sequence of hints that 
guides the reader towards a formal statement in Lean.

Guidelines:
Do not include any Lean code.
Hints must be succinct and make use of mathematical notation.
Do not include proof steps—ignore any part that concerns only the proof.
Ensure that all variables, functions, and assumptions are clearly introduced 
and well-defined.
Use the hints to bridge the gap between the worded (informal) problem and 
the underlying mathematics—make clear how each mathematical concept 
corresponds to elements of the informal statement.
Refer to the following examples of previously generated hints for style 
and structure.

{
EXAMPLE {{ example.id }}:

**Informal statement**
{{ example.informal_statement }}

**Hints**
{{ example.cot}}

{

**Informal statement**
{{ query.informal_statement }}

**Hints**
\end{verbnobox}
\\ \bottomrule

\end{tabular}
\caption{Jinja templates for the system and user prompt used in LeanFIRE for the generation of informal reasoning steps (CoT).}
    \label{fig:prompt-cot}
\end{figure}
\begin{figure}[t!]
\centering
\footnotesize
\begin{tabular}{p{13cm}}
\paragraph{Lean-of-Thought (LoT) Translation Prompt}
\\
\toprule
\textbf{System Prompt} \\

\begin{verbnobox}[\fontsize{8pt}{8pt}\selectfont]
You are an advanced assistant specializing in formal mathematics and Lean 4 
theorem proving. You have extensive expertise in translating mathematical 
concepts from natural language into precise Lean 4 code.

\end{verbnobox}

 \\ \midrule

\textbf{User Prompt} \\

\begin{verbnobox}[\fontsize{8pt}{8pt}\selectfont]
Using the provided hints, write a Lean4 code snippets for each hints when 
appropriate to guide the reader towards a formal statement in Lean.

Guidelines:
Do not provide formal proofs or imports.
Ensure that you match the hints to the Lean hints.
Refer to the following examples of previously generated hints for style 
and structure.

{
EXAMPLE {{ example.id }}:

**Informal statement**
{{ example.informal_statement }}

**Hints**
{{ example.cot}}

**Lean Hints**
{{ example.lot}}

{

**Informal statement**
{{ query.informal_statement }}

**Hints**
{{ example.cot}}

**Lean Hints**
\end{verbnobox}
\\ \bottomrule

\end{tabular}
\caption{Jinja templates for the system and user prompt used in LeanFIRE for the translation of the CoT into formal reasoning steps (LoT).}
\label{fig:prompt-lot}
\vspace{0cm}
\end{figure}

\newpage
\section{Details on Experimental Setup}\label{app: experiments}
This Appendix provides details on our experimental setup. All experiments were conducted in Lean \texttt{v4.19.0-rc2} with the appropriate Mathlib imports and standard LLM APIs for \gpt and \deepseek. Each instance was run for 10 passes using the random seeds [5049, 891, 1065, 4894, 3277, 8476, 8192, 688, 377, 3568] to ensure reproducibility. The only non-default generation parameter was a temperature of 0.7; all other settings were kept at their default values. Prompts for autoformalisation, conjecture generation, and \ConjectureJudge are provided in Sections~\ref{app:autoformalisation-instructions}, \ref{app:conjecturing-instructions}, and \ref{app: ConJudge}, respectively. The Lean 4 code for \texttt{equiv\_rfl} is included in Section~\ref{app: equiv_rfl}.

\newpage
\subsection{Autoformalisation Prompt} \label{app:autoformalisation-instructions}
\begin{figure}[h!]
\centering
\footnotesize
\begin{tabular}{p{13cm}}
\paragraph{Autoformalisation Prompt}
\\
\toprule
\textbf{System Prompt} \\

\begin{verbnobox}[\fontsize{8pt}{8pt}\selectfont]
You are an advanced assistant specializing in formal mathematics and Lean 4 
theorem proving. You have extensive expertise in translating mathematical 
concepts from natural language into precise Lean 4 code.

\end{verbnobox}

 \\ \midrule

\textbf{User Prompt} \\

\begin{verbnobox}[\fontsize{8pt}{8pt}\selectfont]
Translate the following natural language statement, provided under 
**Informal statement** into a formal Lean 4 theorem. Use the theorem name 
specified under **Name** as the Lean identifier for the theorem. Your 
response must:
- Write only valid Lean 4 code, with clear and idiomatic use of Lean 
syntax and conventions.
- Only include the formalization, and do not include any proof or imports.
- Define the theorem using the provided name.
- Faithfully capture the meaning of the informal statement in your 
formalization.
- Enclose all Lean code within triple backticks

Output:
```lean
theorem [NAME] : [Lean formalization of the statement] := sorry
```

{
EXAMPLE {{ example.id }}:

**Name**
{{ example.name }}

**Informal statement**
{{ example.informal_statement }}

The code below presents a solution implementation written in Lean 4. 
This solution has already been incorporated into the current Lean 
environment and is available for use in the formalization.

import Mathlib
{
{{ example.conjecture }}
{

Output:
```lean
{{ example.formal_statement }}
```

Above are examples for you to model the translation of the below natural 
language statement into a Lean 4 formal theorem:

{

**Name**
{{ query.name }}

**Informal statement**
{{ query.informal_statement }}

The code below presents a solution implementation written in Lean 4. 
This solution has already been incorporated into the current Lean 
environment and is available for use in the formalization.

import Mathlib
{
{{ example.conjecture }}
{

**Combined Hints**
{{ query.combined_cot_lot }}

Output:
```lean

\end{verbnobox}
\\ \bottomrule

\end{tabular}
\caption{Jinja templates for the system and user prompt for autoformalisation.}
    \label{fig:prompt-autoform}
\end{figure}

\clearpage
\subsection{Standalone Conjecture Generation Prompt}\label{app:conjecturing-instructions}
\begin{figure}[h!]
\centering
\footnotesize
\begin{tabular}{p{13cm}}
\paragraph{Conjecturing Prompt}
\\
\toprule
\textbf{System Prompt} \\
\begin{verbnobox}[\fontsize{8pt}{8pt}\selectfont]
You are an advanced assistant specializing in formal mathematics and Lean 4 
theorem proving. You have extensive expertise in translating mathematical 
concepts from natural language into precise Lean 4 code. 

You do not provide proofs or full theorem statements, only the mathematical 
expression representing the solution, proposition, or the value being asserted. 
You should first analyze the informal problem statement, then provide the final 
expression as valid Lean 4 code.
\end{verbnobox}

\\ \midrule

\textbf{User Prompt} \\

\begin{verbnobox}[\fontsize{8pt}{8pt}\selectfont]
Your task is to take a natural language mathematical statement and extract the 
mathematical expression, proposition, or value, representing it as a Lean 4 
expression.

**Instructions:**
1. Analyze the informal problem statement to deconstruct its mathematical components.
2. Provide the final solution as a single Lean 4 expression.
3. Present the final output inside a Lean code block, using:
```lean
abbrev solution {solution code}
```

**Informal statement**
{{ example.informal_statement }}

\end{verbnobox}
\\ \bottomrule

\end{tabular}
\caption{Jinja template for the system and user prompt used in to generate a conjecture in Lean 4.}
\label{fig:prompt-extraction}
\end{figure}

\subsection{\ConjectureJudge} \label{app: ConJudge}
\ConjectureJudge evaluates whether a conjecture appears in a given formalised statement. We first conducted human annotations to identify which model and prompt best align with human judgments; this model was then selected as our LLM-as-a-judge. Table~\ref{tab:true-false-counts-wide} presents the distribution of human annotations for 100 sample generations, while Table~\ref{tab:model-scores-wide} reports the accuracy of four different models against the human gold labels. The prompt used for \ConjectureJudge is provided below.

\begin{table*}[h]
\begin{center}
{
\def\arraystretch{1.1}
\setlength{\tabcolsep}{4pt}
\fontsize{8pt}{9pt}\selectfont
\begin{tabularx}{0.3\linewidth}{l X X X}
\toprule
 & \textbf{TRUE} & \textbf{FALSE} & \textbf{Total} \\
\midrule
\textbf{Seen}   & 35 & 11 & 46 \\
\textbf{Unseen} & 21 & 33 & 54 \\
\midrule
\textbf{Total}  & 56 & 44 & 100 \\
\bottomrule
\end{tabularx}
}
\caption{Contingency table showing counts of TRUE and FALSE values for seen and unseen instances.}
\label{tab:true-false-counts-wide}
\end{center}
\end{table*}

\begin{table*}[h!]
\begin{center}
{
\def\arraystretch{1.1}
\setlength{\tabcolsep}{6pt}
\fontsize{8pt}{9pt}\selectfont
\begin{tabularx}{0.35\linewidth}{l X}
\toprule
\textbf{Model} & \textbf{Percentage} \\
\midrule
internlm2-math-plus-20b       & 60 \\
qwen3-14b                      & 79 \\
gpt-oss-20b                   & 70 \\
qwen3-30b-a3b-instruct        & \textbf{83} \\
\bottomrule
\end{tabularx}
}
\caption{Percentage alignment to human annotators for \ConjectureBench across different models.}
\label{tab:model-scores-wide}
\end{center}
\end{table*}

\newpage
\begin{figure}[t]
\centering
\footnotesize
\begin{tabular}{p{13cm}}
\paragraph{\ConjectureJudge Evaluation Prompt}
\\
\toprule
\textbf{System Prompt} \\
\begin{verbnobox}[\fontsize{8pt}{8pt}\selectfont]

You are an expert in the Lean 4 theorem proving language and formal 
mathematics. Your task is to determine if a given formal statement in 
Lean 4 contains a specific conjectured value, algebraic formula, or bound.

You will be given three inputs:
1. **Conjecture**: The value, formula, or bound to look for.
2. **Ground Truth Formal Statement**: An example of a Lean 4 statement that 
correctly formalizes the conjecture. Use this as a reference for a valid 
implementation.
3. **Formal Statement**: The Lean 4 code you need to evaluate.

Your goal is to determine if the **Formal Statement** contains the core 
assertion of the **Conjecture**. The **Ground Truth Formal Statement** is 
provided to help you understand how the conjecture can be formally expressed.

The statement you are evaluating might not have the exact same syntax as the 
ground truth. You must carefully check for **semantically equivalent 
variations** of the conjecture's core idea. This includes, but is not limited 
to, permutations of terms, different but equivalent algebraic expressions, or 
reordered hypotheses. Additionally, a conjecture can be expressed either by 
defining a proposition (e.g., `abbrev conjecture : Prop := ...`) or by 
asserting it within a theorem, which implicitly states the conjecture holds. 
You should consider these forms equivalent.

Your output must follow this structure exactly:
1. First, provide a brief explanation of your reasoning.
2. Second, conclude with the final answer in the format: 'The formal 
statement contains the conjecture: **True**' or 'The formal statement 
contains the conjecture: **False**'.

\end{verbnobox}
\\
\midrule

\textbf{User Prompt} \\

\begin{verbnobox}[\fontsize{8pt}{8pt}\selectfont]
**Conjecture:**
```lean
{{ conjecture }}
```

**Ground Truth Formal Statement:**
```lean
{{ statement1 }}
```

**Formal Statement:**
```lean
{{ statement2 }}
```
\end{verbnobox}
\\
\bottomrule

\end{tabular}
\caption{Jinja templates for the system and user prompts used by \textsc{ConJudge}.}
\label{fig:prompt-conjudge}
\end{figure}
\subsection{\texttt{equiv\_rfl}}\label{app: equiv_rfl}
\begin{figure}[h!]
\begin{myleanbox}
\begin{lstlisting}
abbrev conjecture_gold: {gold}
abbrev conjecture_generated: {generated}

theorem thm : conjecture_gold = conjecture_generated := by rfl
\end{lstlisting}
\end{myleanbox}
\caption{Implementation of metric \texttt{equiv\_rfl} in Lean 4.}
\label{fig:equivrfl-code}
\end{figure}

\end{document}